\documentclass[letterpaper]{article} 
\usepackage{aaai2026}  
\usepackage{times}  
\usepackage{helvet}  
\usepackage{courier}  
\usepackage[hyphens]{url}  
\usepackage{graphicx} 
\usepackage{natbib}  
\usepackage{caption} 
\usepackage{algorithm}
\usepackage{algorithmic}
\usepackage[utf8]{inputenc}
\usepackage{subcaption}
\usepackage{lipsum}

\usepackage{newfloat}
\usepackage{listings}
\DeclareCaptionStyle{ruled}{labelfont=normalfont,labelsep=colon,strut=off} 
\floatstyle{ruled}
\newfloat{listing}{tb}{lst}{}
\floatname{listing}{Listing}

\usepackage{xcolor}

\usepackage{multirow}
\usepackage{amsmath}

\title{SlideTailor: Personalized Presentation Slide Generation for Scientific Papers}
\author{
    Wenzheng Zeng\equalcontrib,
    Mingyu Ouyang\equalcontrib,
    Langyuan Cui\equalcontrib,
    Hwee Tou Ng\thanks{Corresponding author.}
}
\affiliations{
    Department of Computer Science, National University of Singapore\\
    \{wenzhengzeng, ouyangmingyu04, langyuan.c\}@u.nus.edu,
    dcsnght@nus.edu.sg
    
}

\usepackage{bibentry}

\begin{document}

\maketitle

\begin{abstract}
Automatic presentation slide generation can greatly streamline content creation. However, since preferences of each user may vary, existing under-specified formulations often lead to suboptimal results that fail to align with individual user needs. We introduce a novel task that conditions paper-to-slides generation on user-specified preferences. We propose a human behavior-inspired agentic framework, SlideTailor, that progressively generates editable slides in a user-aligned manner. Instead of requiring users to write their preferences in detailed textual form, our system only asks for a paper-slides example pair and a visual template—natural and easy-to-provide artifacts that implicitly encode rich user preferences across content and visual style. Despite the implicit and unlabeled nature of these inputs, our framework effectively distills and generalizes the preferences to guide customized slide generation. We also introduce a novel chain-of-speech mechanism to align slide content with planned oral narration. Such a design significantly enhances the quality of generated slides and enables downstream applications like video presentations. To support this new task, we construct a benchmark dataset that captures diverse user preferences, with carefully designed interpretable metrics for robust evaluation. Extensive experiments demonstrate the effectiveness of our framework.

\end{abstract}

\begin{links}
    \link{Project website}{https://github.com/nusnlp/SlideTailor}
\end{links}

\section{Introduction}
Presentations, usually delivered through slides, are a widely used medium for effectively communicating information in a visually engaging and accessible way~\cite{effectiveness_of_ppt_2003}. Crafting high-quality presentations, however, demands considerable effort, requiring the author to put in informative and focused content, craft a coherent and compelling narrative, and create an appealing visual design. Given the time and expertise required, there is a growing interest in developing automated systems that can generate presentation slides to reduce the manual workload involved.

Recent works~\cite{doc2ppt_aaai22, pptagent, t2v_iterative_arxiv25} have exploited the inherent multimodal nature of the academic document-to-slides generation task, showing promising results in both content quality and visual layout.
Despite the effectiveness, they typically treat slide generation as a straightforward document-to-slides conversion task. This overlooks a crucial aspect: the user. We argue that presentation design is inherently subjective. 
Users have different preferences in terms of narrative structure, emphasis, conciseness, and aesthetic choices. Consequently, under-specified or one-size-fits-all generation frameworks often yield outputs that do not align well with individual needs. To enable more personalized, user-aligned presentations, it is essential to incorporate individual preferences.

\begin{figure}[t] 
    \centering
    \includegraphics[width=1.0\linewidth]{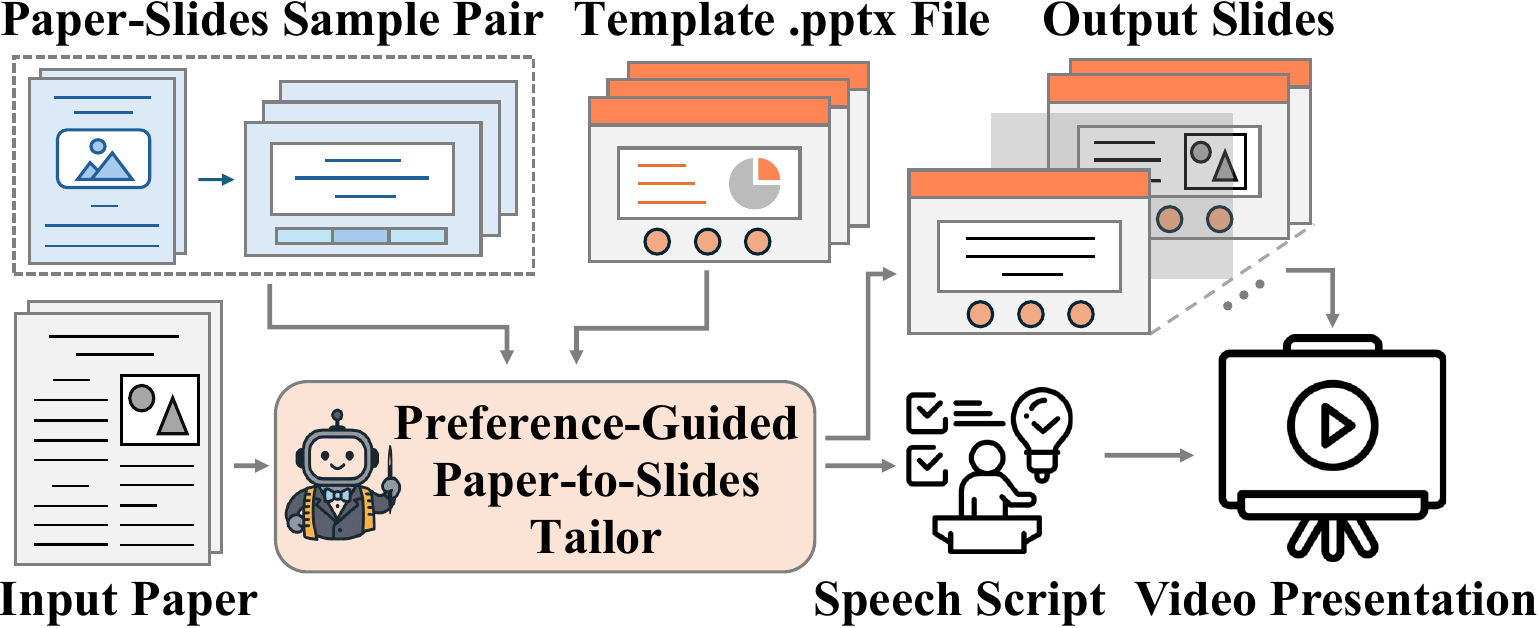}
    \caption{Preference-guided paper-to-slides generation. Based on user preferences inferred from a paper-slides sample pair and a visual template, the system produces personalized slides accompanied by a speech script, supporting downstream applications such as video presentations.}
    \label{fig:fig1}
\end{figure}

Motivated by this, we shed light on this under-explored research problem: preference-guided paper-to-slides generation, where the generation process is explicitly guided by user-specified preferences. We categorize such preferences into two main aspects: (1) Content preferences, which affect narrative flow and the level of emphasis or conciseness applied to specific topics or sections; and (2) Aesthetic preferences, which govern layout structure, background design, decorative elements, and overall stylistic choices.

Instead of asking users to articulate their preferences in detailed textual instructions, we propose a more user-friendly way. The system takes (1) a paper-slides sample pair, which implicitly encodes the user’s content structuring-related preferences, and (2) a \texttt{.pptx} slide template, which reflects aesthetic choices. These inputs are natural to provide and align with how users often prepare slides—by referencing prior slide decks and reusing templates. Besides, these two types of inputs are relatively orthogonal, offering greater clarity and flexibility by enabling customized slide generation along separate content and aesthetic dimensions. However, while these inputs are convenient for users to provide, they pose nontrivial challenges for the slide generation model: the embedded preferences are implicit, entangled, and unlabeled, making them difficult for the model to extract and apply effectively.

To tackle this challenge, we propose a human behavior-inspired agentic framework termed SlideTailor. It progressively constructs editable slides aligned with user preferences. The process begins with preference distillation, similar to a human that summarizes and learns multi-aspect user preferences from both the given sample pair and \texttt{.pptx} template file, forming an internalized preference profile. Guided by this profile, the system performs a preference-guided and presentation-oriented summarization process. It extracts and reorganizes salient content from the input paper, while adjusting the level of detail, emphasis, and narrative flow to align with the user’s preferred presentation style. The resulting content is then structured into a coherent outline across slides, specifying the intended message and supporting points for each slide.

As a novel component of our framework, we introduce a chain-of-speech mechanism during the outline construction. Inspired by how human presenters plan their speech alongside slide design, this mechanism prompts the system to simulate narrative when outlining each slide. As a result, slide content can better align with the anticipated speech, improving coherence and clarity. It also enables downstream applications such as full video presentations (Fig.~\ref{fig:fig1}).

Based on the constructed outline, the system proceeds to template planning, selecting the most appropriate layout for each slide based on its semantic content and intended visual emphasis. This step ensures that the slide structure and aesthetics are jointly optimized in line with user preferences.
Finally, the system generates slides by editing the selected templates and exporting them in standard \texttt{.pptx} format, which enables flexible user refinement and downstream use.

To facilitate research on this new task, we construct a benchmark dataset that captures and simulates diverse user preferences to comprehensively evaluate customized paper-to-slides generation methods. We also carefully craft interpretable metrics for robust evaluation of preference alignment and presentation quality. Experimental results demonstrate that our method not only better aligns with user intent, but also produces slides with higher overall quality compared to existing approaches.

\section{Related Work}
\subsection{Document-to-Slides Generation}
Prior works primarily considered slide generation as a text summarization problem~\cite{topic_aware_pure_text_aaai21, d2s_naacl21, text_no_condition_2023, not_linear_emnlp_pure_text_finding24, knowledge_centric_pure_text_arxiv24}, overlooking layout design and the inherent multimodal nature of engaging presentations. Some works~\cite{doc-to-poster-gen_arxiv2021, doc2ppt_aaai22, enhancing_presentation_inlg24}, especially recent and contemporaneous studies~\cite{pptagent, paper2poster, presentagent, paper2video}, began integrating visual and layout elements for multimodal presentation. Despite their effectiveness, most methods treat slide generation as a direct document-to-slides process, lacking constraints that capture diverse user preferences. Persona-Aware-D2S~\cite{persona_aware_d2s_eacl24} considers customization, but restricts preferences to four fixed categories. PPTAgent~\cite{pptagent} enables flexible template input, but focuses solely on layout aspects, overlooking content-related preferences.
In contrast, we explore the subjective nature of slide generation in a more realistic formulation, with essential contributions on definition, methodology, and dataset, aiming to facilitate future research on personalized slide generation.

\subsection{Conditional Summarization}
This task generates summaries conditioned on auxiliary inputs beyond the source itself, such as queries~\cite{query_based_summarization_survey, lmgqs_query_based_summarization_bart, ideal_query_based_summarization_llm}, topic cues~\cite{topic_aware_pure_text_aaai21, topic_aware_multimodal_summarization}, timeline~\cite{timeline_summarization_acl24, timeline_summarization_aaai25}, diagram~\cite{scidoc2diagrammer_emnlpfindings24}, and user preferences~\cite{personalized_review_summarization}. The subarea most relevant to our work concerns user preferences. Within this subarea, different studies focus on different facets—for example, review summarization with personalized recommendations~\cite{personalized_review_sum_aai219, personalized_review_sum_arxiv23, personalized_review_summarization, sumrecom_arxiv24} and controllable abstractive summarization where users specify attributes like style or length~\cite{Controllable_abstractive_summarization}. 
In contrast to these tasks, we focus on the conditional summarization of scientific papers for presentation slide generation. The work most similar to ours is~\cite{persona_aware_d2s_eacl24}. However, it is limited to four predefined preferences (i.e., expert/non-expert, long/short), which fail to capture the diverse and fine-grained nature of real user needs. Our formulation instead models diverse and fine-grained user preferences in a more realistic, open-ended setting.

\begin{figure*}[t] 
    \centering
    \includegraphics[width=0.9\textwidth]{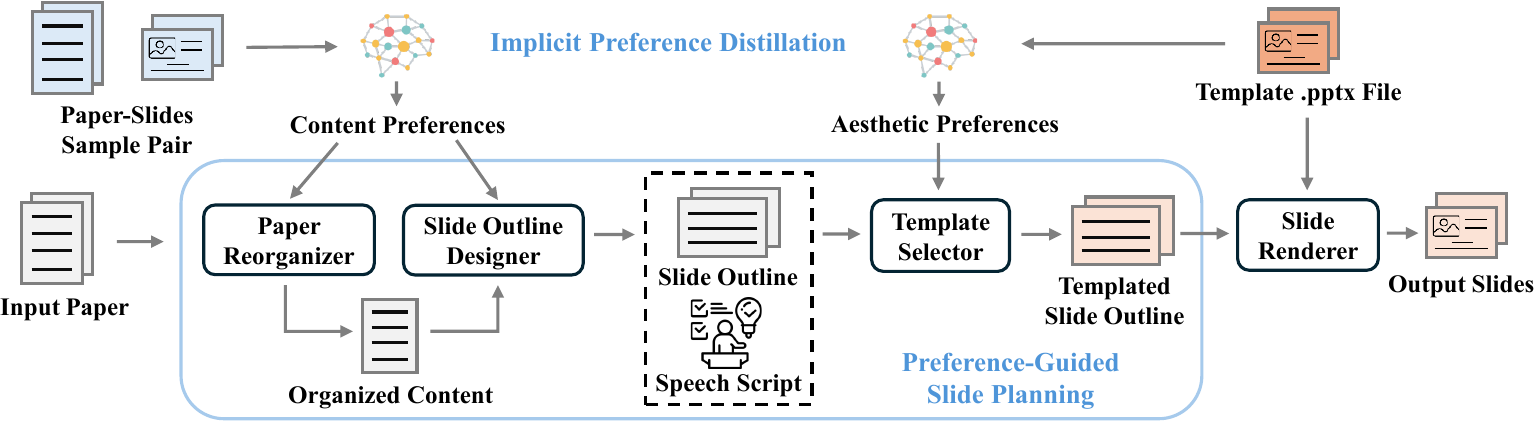}
    \caption{The conceptual pipeline of our proposed preference-guided paper-to-sildes generation framework.}
    \label{fig:pipeline}
\end{figure*}

\section{Task Formulation} \label{sec: task}

We now formulate the preference-guided paper-to-slides generation task.
We begin by defining the key components that constitute user preferences in the context of presentation slide generation. We distinguish two preference dimensions:
(1) Content preferences, which shape how a paper is mapped into presentation—affecting the narrative flow, level of detail, emphasis placed on certain sections (e.g., highlighting results over methodology), and decisions about what content to omit or elaborate;
(2) Aesthetic preferences, which reflect the user’s visual and stylistic inclinations—governing choices such as slide layout (e.g., text-only vs. text-image combination), background themes, color palettes, typography, and the use of decorative elements like icons or logos.

Next, we explore what forms of user input can both effectively capture expressive user preferences and align well with real-world presentation creation workflows. 
Rather than requiring users to explicitly articulate their preferences through detailed textual instructions—which can be unnatural and burdensome—we adopt a more user-friendly and example-driven setting. Specifically, the system takes two types of input:
(1) a \emph{paper-slides sample pair}, which captures the content preferences; and
(2) a \texttt{.pptx} \emph{slide template}, which conveys aesthetic preferences such as layout style and visual theme.
These inputs are natural for users to provide and mirror common practices in presentation authoring, where individuals often reference previous slide decks and reuse institutional or personal templates. 
Moreover, the two preference types are relatively orthogonal, enabling modular flexibility—any aesthetic template can, in principle, pair with any content preference profile. In practice, we assume users have a general sense of their desired outcomes, yielding self-consistent preferences, while the designed model is expected to exhibit adaptive capability to achieve preference harmony across different preference sources during slide generation.

Formally, given a paper $D$ in PDF format, the goal is to synthesize the corresponding slides $S$ that (i) satisfy the general quality requirements of slide production from $D$, and (ii) follow the subjective preference embodied by the aforementioned two types of conditional user inputs: a paper-slides sample pair $(D_{\text{ref}}, S_{\text{ref}})$ for content preferences, and a template \texttt{.pptx} file $S_{\text{tmpl}}$ for aesthetic preferences. This can be formally defined as:
\begin{equation}
    S = F\left(D, (D_{\text{ref}}, S_{\text{ref}}), S_{\text{tmpl}}\right),
\end{equation}
where $F$ is the function that jointly performs conditional paper summarization and slide generation.

\section{Method}
In this section, we introduce our proposed framework SlideTailor for personalized paper-to-slides generation. We begin by analyzing the core challenges of this newly introduced task, which in turn motivate our design. We then present the framework's components.

\subsection{Key Challenges}
While the aforementioned setup in Sec.~\ref{sec: task} eases the burden on users and better reproduces real-world authoring practice, it poses two key challenges:
\paragraph{C\textsubscript{1}: Learning from implicit, unlabelled preference signals.} Although the user-supplied sample pair $(D_{\text{ref}}, S_{\text{ref}})$ and template $S_{\text{tmpl}}$ provide rich preference signals, they are implicit and entangled without structured labels. Thus, the system must distill diverse content and aesthetic preferences without fine-grained supervision, making preference extraction inherently ambiguous and under-constrained.
\paragraph{C\textsubscript{2}: Multi-aspect alignment.} 
Guided by the distilled preferences, slide generation should achieve harmonious alignment across content and aesthetic dimensions.

\medskip\noindent To address C\textsubscript{1} and C\textsubscript{2}, we introduce a human behavior-inspired agentic framework SlideTailor that progressively models and applies user preferences in slide generation. 
As illustrated in Fig.~\ref{fig:pipeline}, the overall process involves three stages: (1) implicit preference distillation from unlabeled user inputs, (2) preference-guided slide planning from the distilled profile, and (3) slide realization via appropriate template editing. This modular design mirrors slide preparation by humans: starting from internalizing personal style, then organizing content, and finally creating slides for presentation.

\subsection{Implicit Preference Distillation}

The first stage extracts user preferences from two types of example-driven inputs: a paper-slides sample pair $(D_{\text{ref}}, S_{\text{ref}})$, and a template \texttt{.pptx} file $S_{\text{tmpl}}$. These inputs are unlabeled, with no explicit annotation of what the user prefers, but still carry rich preference cues across content structure and visual design. We approach this preference inference by designing a dual-branch distillation process that aims at converting them into explicit, structured, and interpretable preferences for subsequent slide generation. Our key insight is to treat the sample pair $(D_{\text{ref}}, S_{\text{ref}})$ as an implicit demonstration of how the user transforms source content into a presentation, while treating the template $S_{\text{tmpl}}$ as a reflection of their aesthetic inclinations. These two sources are complementary: the former governs \emph{what} and \emph{how} to present; the latter specifies \emph{how it should look}.

\begin{figure}[t] 
    \centering
    \includegraphics[width=0.97\linewidth]{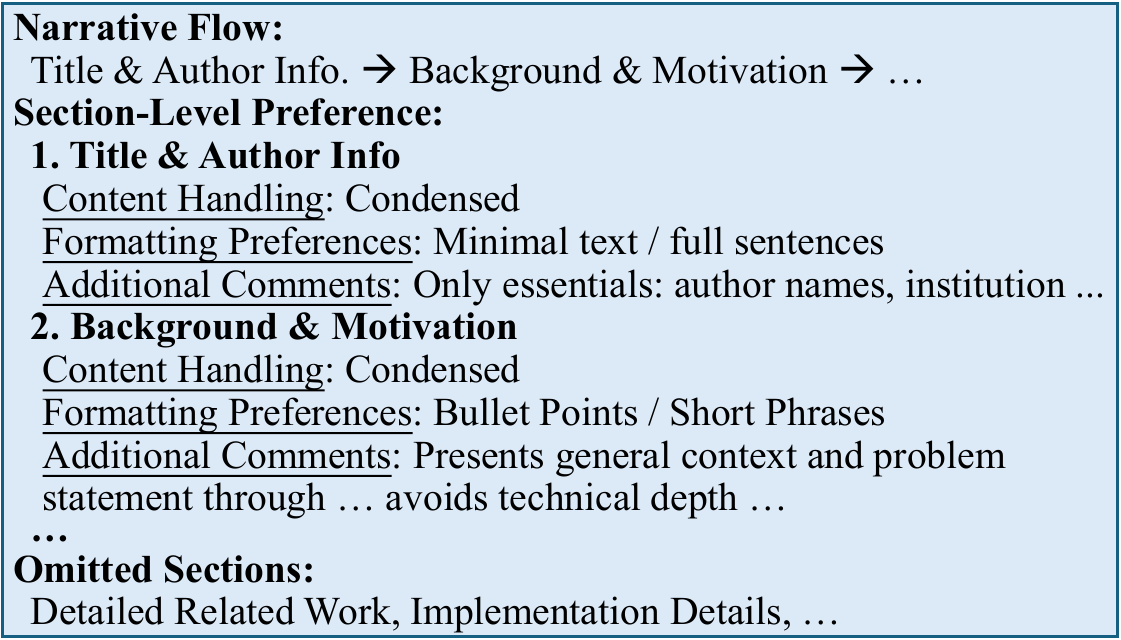}
    \caption{An example of distilled content preference.}
    \label{fig:content_preference}
\end{figure}

\noindent \paragraph{Content preference as a latent mapping.}
~We model the transformation $f_{\text{content}}: D_{\text{ref}} \rightarrow S_{\text{ref}}$ as a latent function that captures the user’s personal preferences for abstraction and organization. Instead of aligning exact sentences or keywords, we leverage a large language model (e.g., GPT-4.1~\cite{gpt41}) to infer how the content is selected, emphasized, omitted, or reordered, beyond surface form. This yields a structured content preference profile $P_C$ comprising: (1) narrative flow (e.g., Title → Background \& Motivation → ... → Future Work), and (2) section-level emphasis or omission, as well as formatting preferences (e.g., use of bullet points). The LLM is also encouraged to provide additional comments when deemed helpful, allowing the extracted structure to remain flexible and adaptive to context. An illustration example is shown in Fig.~\ref{fig:content_preference}.

\begin{figure}[t] 
    \centering
    \includegraphics[width=0.97\linewidth]{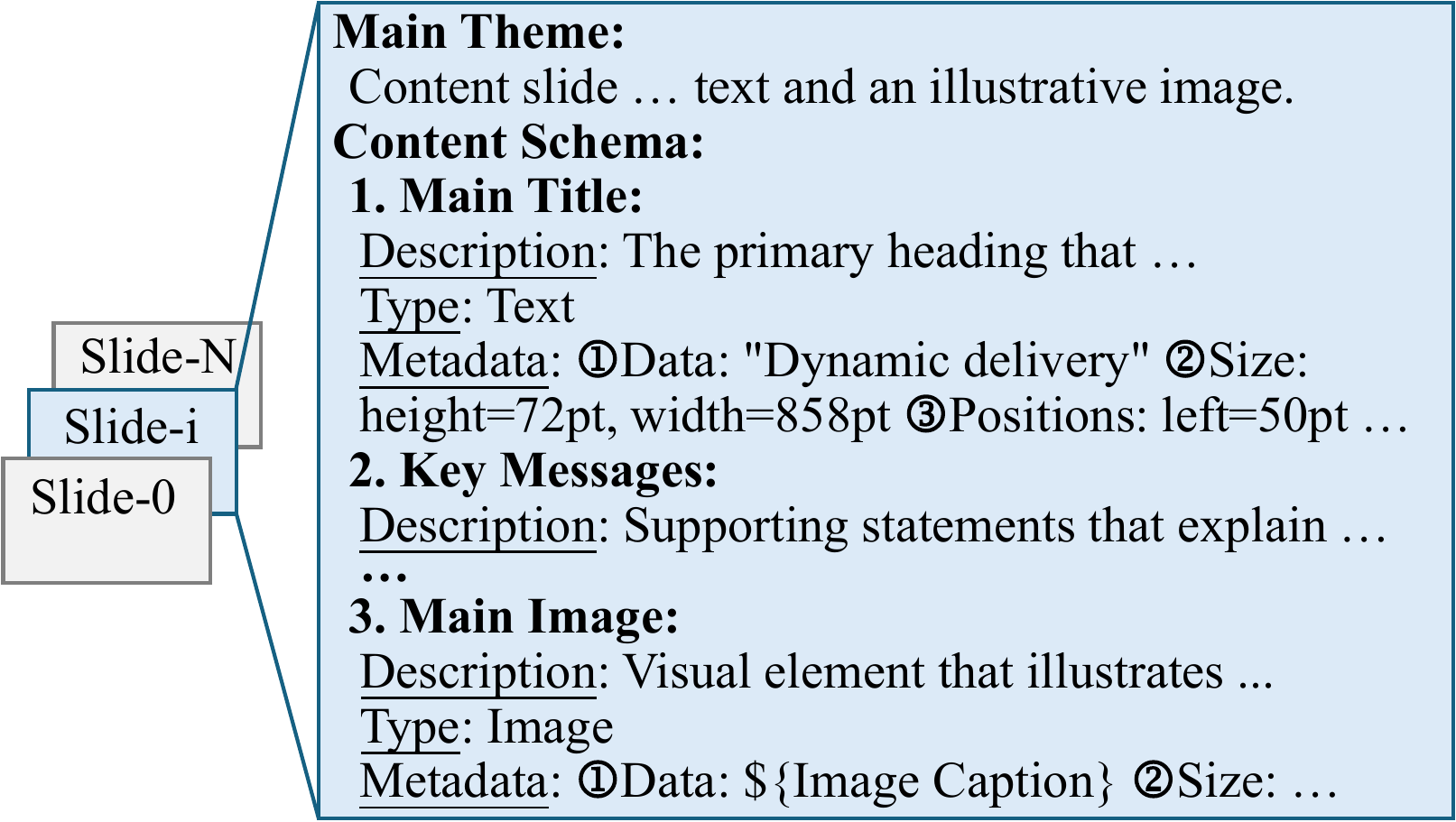}
    \caption{An example of distilled aesthetic preference.}
    \label{fig:aesthetic_preference}
\end{figure}

\noindent \paragraph{Aesthetic preference from the template.}
~In parallel, we distill layout-related aesthetic preferences from the provided template $S_{\text{tmpl}}$. We employ a vision-language model (VLM) to infer the functional roles of both slide-level components (e.g., title, main content, conclusion) and element-level components (e.g., text boxes, image regions) within each slide.
In addition, we incorporate fine-grained metadata directly parsed from the raw \texttt{.pptx} file, such as precise bounding box/image positions and sizes.
The resulting structured schema $P_A$, as shown in Fig.~\ref{fig:aesthetic_preference}, serves as a layout grounding mechanism that facilitates subsequent template understanding and selection.

Finally, the union $P = P_C \cup P_A$ constitutes a modular, symbolic representation of user preferences. Subsequent stages use $P$ as the conditioning context, ensuring that the generated slides simultaneously reflect the sample’s structural style and the template’s aesthetics.

\begin{figure*}[t] 
    \centering
    \includegraphics[width=0.99\textwidth]{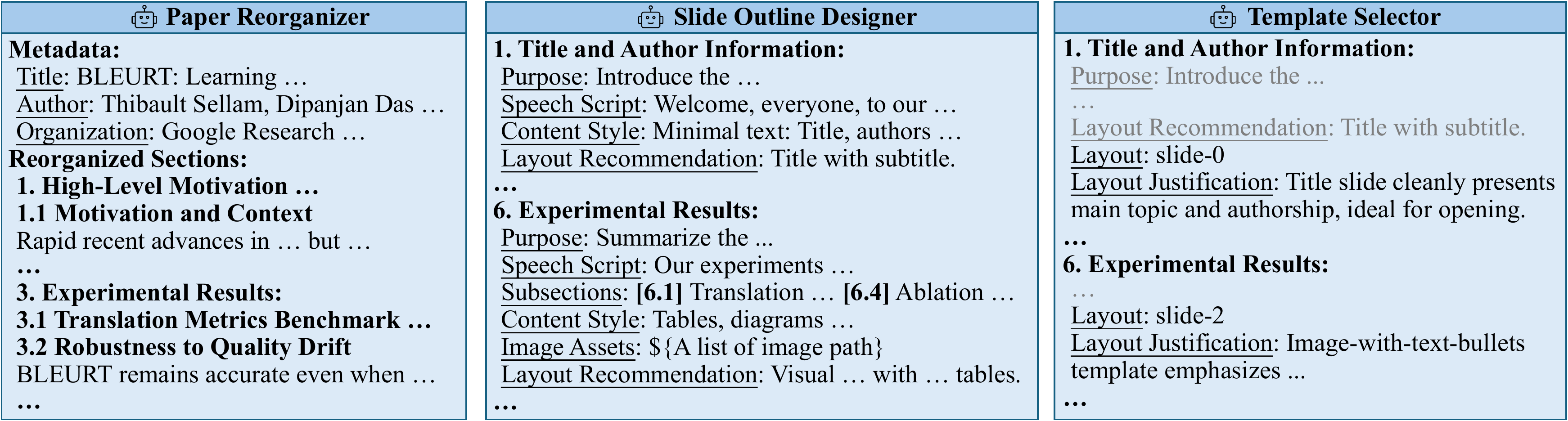}
    \caption{Example output from paper reorganizer (left), slide outline designer (middle), and template selector (right).}
    \label{fig:3_llm_agent_output}
\end{figure*}

\subsection{Preference-Guided Slide Planning}

Guided by distilled preferences $P$, the system progressively plans the presentation by deciding what to say, structuring it into slides, and assigning visual formats. This process is handled by three LLM-powered agents: a paper reorganizer that restructures an input paper based on preferences, a slide outline designer that segments content into slides and drafts narration, and a template selector that assigns a suitable template for each planned slide.

\noindent \textbf{Conditional paper reorganizer.} 
Unlike generic summarization, our LLM agent reorganizes the input paper to reflect user-specific priorities encoded in $P_C$—adjusting the narrative flow and the level of detail based on presentation preferences. The result is a presentation-oriented document that forms the content for subsequent slide generation, as shown in the left part of Fig.~\ref{fig:3_llm_agent_output}.

\noindent \textbf{Slide-wise outline generation with chain-of-speech.}
Next, the reorganized content is segmented into a coherent slide-wise outline, where each slide plan specifies the intended message and visual cues (e.g., images, tables from the paper). Inspired by how presenters rehearse during slide creation, we introduce a \textit{chain-of-speech} mechanism that simultaneously drafts speech for each slide. This encourages alignment between visual content and anticipated oral delivery, improving the coherence and usability of the final output. The resulting slide outline and speech script (as shown in the middle part of Fig.~\ref{fig:3_llm_agent_output}) will jointly guide both content realization and layout selection. This design also facilitates downstream applications, as illustrated in Sec.~\ref{sec:downstream}.

\noindent \textbf{Template-aware layout selection.}
Once the content outline is finalized, the system selects an appropriate layout for each slide by matching it with one slide from the user-provided template, based on the slide-level aesthetic schema $P_A$, as shown in the right part of Fig.~\ref{fig:3_llm_agent_output}. This per-slide matching aligns with real-world authoring practices and enhances coherence between slide content and visual presentation.

\subsection{Slide Realization}
Finally, the system realizes each slide by editing the selected template layout using the outline and corresponding speech draft. A layout-aware agent maps planned content (e.g., titles, text, visuals) to specific elements (e.g., text boxes, image placeholders) in the assigned template. This structured mapping may involve modifying, replacing, or inserting elements for coherence. A code agent then generates executable code to apply these edits directly to the \texttt{.pptx} file. This editing-based strategy preserves the original layout and theme while producing fully editable slides suitable for further user refinement.

\subsection{Downstream: Video Presentations} \label{sec:downstream}
Beyond slide generation, our framework could potentially support downstream applications such as automated, speaker-aware video presentations. Thanks to the proposed chain-of-speech mechanism, each slide is paired with a generated speech script $T$, which can be directly transformed into personalized narration using existing zero-shot text-to-speech systems~\cite{openvoice, megatts3}. With just a short voice sample, the synthesized speech could preserve the user's vocal identity~\cite{openvoice}. Combined with the visual slides $S$, this could enable the automatic creation of synchronized, personalized presentation videos, offering a scalable solution for remote teaching or pre-recorded conference talks.
In addition, other compelling extensions can be integrated. For example, an identity-preserving talking head can be synthesized using existing audio-driven generation methods~\cite{memo, actalker} and incorporated into the video to further enhance realism and audience engagement. In this work, we also take a step toward realizing this downstream extension. The implementation details can be found in our project website.

\begin{table*}[t]
\setlength{\tabcolsep}{4pt} 
\centering
\scriptsize
\begin{tabular}{l|ccc|c|c}
\hline
Benchmark                 & \# Test Papers & \# Preference Types & \# Combinations & Open-Ended Preference?    & Source Domain                                                                  \\ \hline
SciDuet~\cite{d2s_naacl21}                   & 81             & -                   & 81              & -                            & 3 AI Conferences                                                        \\
DOC2PPT~\cite{doc2ppt_aaai22}                   & 595            & -                   & 595             & -                            & 9 AI Conferences              \\
Persona-Aware-D2S~\cite{persona_aware_d2s_eacl24}         & 50             & 4                   & 200             & No                           & Subset of SciDuet                                                      \\ \hline
\multirow{2}{*}{PSP (Ours)} & \multirow{2}{*}{200} & \multirow{2}{*}{500} & \multirow{2}{*}{100{,}000} & \multirow{2}{*}{Yes} & 7 AI Conferences, 3 Biomedical Journals, \\
                           &                 &                      &                   &                             & 1 Chemistry Journal, 1 General Journal \\ \hline
\end{tabular}
\caption{Comparison among existing paper-to-slides generation benchmarks.}
\label{tab:data_stat}
\end{table*}

\section{The PSP Dataset}

To facilitate research on the proposed new task, we construct a dedicated benchmark dataset, PSP (\underline{P}aper-to-\underline{S}lides with \underline{P}references), with effort on both data and evaluation protocol. Unlike prior datasets that focus on direct paper-to-slides conversion with limited consideration of user preferences, PSP explicitly incorporates diverse user preferences, covering both content structuring and aesthetic presentation, thereby paving the way for comprehensive evaluation of customized paper-to-slides generation methods.

We manually curated data from the public proceedings of leading AI and scientific venues, including top conferences such as AAAI, ACL, CVPR, NeurIPS, as well as high-impact journals like Nature, Science, The Lancet, and Chemical Reviews Letters. The source corpus encompasses papers spanning a broad spectrum of research fields, including general AI, machine learning, natural language processing, computer vision, chemistry, and medicine, ensuring both topical and stylistic diversity. To capture variation in user presentation preferences, we collected 50 distinct and high-quality paper-slides pairs that reflect diverse structuring and stylistic preferences across presenters and disciplines. Additionally, we curated a set of 10 academic slide templates representative of common research-oriented layout and aesthetic conventions. Finally, we gathered 200 scientific papers to serve as target input papers for slide generation. Altogether, this yields a pool of 200 target papers, 50 sample paper-slides pairs, and 10 layout templates, resulting in up to $200 \times 50 \times 10 = 100{,}000$ unique input combinations for conditional slide generation.
As shown in Table~\ref{tab:data_stat}, our dataset is the largest among existing paper-to-slides generation benchmarks, offering significantly more input combination possibilities and uniquely supporting open-ended preference modeling.

\section{Evaluation Metrics}
To support proper evaluation, we introduce two complementary sets of metrics: preference-based metrics, which assess how well the generated slides can follow user preferences, and preference-independent metrics, which evaluate the overall slide quality independent of those preferences.

\subsubsection{Preference-Based Metrics}
We propose four metrics to evaluate a system's ability to follow user preferences across different aspects.

\noindent \textbf{1. Coverage.}
It evaluates whether the generated slides cover a similar set of high-level subtopics (e.g., introduction, motivation, method) as the sample slides. We use an LLM to extract these structural topics from both $S$ and $S_{ref}$ and compute the intersection-over-union (IoU) between them.

\noindent \textbf{2. Flow.} 
It assesses whether the subtopics are presented in a similar order. Using the same LLM-based extraction method, we obtain the subtopic sequences from both the generated and sample slides, and compute the Normalized General Levenshtein Distance (NGLD)~\cite{ngld} between them. The similarity score is defined as $1 - \text{NGLD}(S, S_{\text{ref}})$, where NGLD lies in $[0, 1]$.

\noindent \textbf{3. Content Structure.}  
We assess how well the generated slides align with the structural presentation style of the input paper–slides sample pair. Using an LLM-as-a-judge framework, the model is instructed to focus on content organization such as pace, level of detail, visual formatting, and slide transitions—while ignoring the actual subject matter. A score from 1 to 5 is assigned based on the degree of structural and stylistic alignment.

\noindent \textbf{4. Aesthetic.}
It evaluates how well the generated slides visually align with the given template, focusing on layout, background, color scheme, fonts, and recurring elements (e.g., headers or logos). The assessment targets visual design only, ignoring textual or semantic content. We feed slide screenshots into a VLM to produce a 1–5 score.

\begin{table*}[t]
 \centering
    \footnotesize
\setlength{\tabcolsep}{8.5pt}
\renewcommand{\arraystretch}{1.1} 
\begin{tabular}{l|cccc|cc|c} \hline
\multirow{2}{*}{Method} & \multicolumn{4}{c}{Preference-based}          & \multicolumn{2}{|c|}{Preference-independent}             & \multirow{2}{*}{Overall} \\
& Coverage      & Flow          & Content Structure         & Aesthetic       & Content    &  Aesthetic &                          \\ \hline
ChatGPT                 & 62.62 & 56.84 & 61.60 & 80.80 & 47.00 & 68.32 & 62.86 \\
AutoPresent (GPT-4.1)   & 72.84 & 59.58 & 49.60 & 22.40 & 28.05 & 60.20 & 48.78 \\ 
PPTAgent (GPT-4.1)      & 64.41 & 54.24 & 57.60 & 97.20 & 58.36 & 71.96 & 67.30 \\ \hline 
Ours (Qwen2.5+Qwen2.5VL)& 70.19 & 62.16 & 68.41 & 92.80 & 48.84 & 72.84 & 69.21 \\
Ours (GPT-4.1)          & \textbf{74.47} & \textbf{66.65} & \textbf{72.80} & \textbf{98.00} & \textbf{67.64} & \textbf{75.24} & \textbf{75.80} \\
\hline        
\end{tabular}
\caption{Performance on the PSP dataset. Comparison of our framework (two backbone variants) against three state-of-the-art baselines. Scores are averaged over 50 target papers.}

    \label{tab:main_experiment}
\end{table*}

\subsubsection{Preference-Independent Metrics}

While capturing user preferences is central to our system, the generated slides should also be of high quality regardless of those preferences. To support a more holistic evaluation, we also introduce a set of metrics that assesses the presentation quality independent of user-specific preferences. Each metric is scored using an MLLM-as-a-judge framework, where the model rates specific aspects of the slides from 1 to 5 following a defined rubric.

\noindent \textbf{1. Content.}
It evaluates how clearly and accurately the slides convey the key information of the target paper. The model considers the relevance and depth of the content and clarity. The goal is to assess whether the slides provide a coherent, focused, and informative summary of the original work.

\noindent \textbf{2. Aesthetic.}
It assesses the overall visual appeal of the generated slides. An MLLM is instructed to focus on the presentation’s design elements—such as layout, color choices, font consistency, visual balance, and spacing—without considering content semantics. The model reviews the slides holistically to determine their professional and aesthetic quality.

Note that all metrics are normalized to a 0–100 scale for consistent comparison across evaluation dimensions. We use GPT-4.1 as the MLLM for all evaluations.
Detailed grading rubrics are provided at our project website.
We also report the average of these metrics, denoted as ``Overall".

\section{Experiments}
\subsection{Experimental Setup}
For the primary evaluation, we randomly sampled 50 papers from the PSP dataset as target input papers. Each paper was independently paired with one paper–slides sample pair and one \texttt{.pptx} template file as preferences, both randomly selected from the benchmark dataset.

We benchmarked our framework against three state-of-the-art slide generation baselines: (1) \textbf{ChatGPT}~\cite{openai2025chatgpt}: We manually upload all input components (target paper, paper-slides sample pair, and template) via its web interface and prompt it to generate slides of the target paper based on the supplied preferences. (2) \textbf{AutoPresent}~\cite{autopresent_cvpr25}: As a text-to-slides generation method, AutoPresent takes only raw text as input. To simulate preference conditioning, we adapted it to our setting by concatenating the plain-text versions of the paper-slides pair with the target paper. (3) \textbf{PPTAgent} \cite{pptagent}: Since PPTAgent does not accept a paper-slides pair as preferences, we only provide the target paper and template as input. For all methods (including ours), we constrain the generation to 10 slides by embedding the instruction into the prompt for a fair comparison.

All compared systems were evaluated in a zero-shot setting powered by MLLMs. Unless otherwise noted, we employed and evaluated each system with the proprietary GPT-4.1 (checkpoint \texttt{gpt-4.1-2025-04-14}) serving as both vision and language model. We also instantiated and evaluated our system with the open-source Qwen2.5-72B-Instruct~\cite{qwen2.5} and Qwen2.5-VL-72B-Instruct models~\cite{Qwen2.5-VL} to demonstrate its adaptability and robustness across base LMs. These open-source models were served through the LMDeploy~\cite{2023lmdeploy} framework and ran on NVIDIA H200 GPUs.

\begin{table*}[t]
 \centering
    \footnotesize
\setlength{\tabcolsep}{10pt}
\renewcommand{\arraystretch}{1.1} 
\begin{tabular}{l|cccc|cc|c} \hline
\multirow{2}{*}{Setting} & \multicolumn{4}{c}{Preference-based} & \multicolumn{2}{|c|}{Preference-independent} & \multirow{2}{*}{Overall} \\
& Coverage & Flow & Content Structure & Aesthetic & Content & Aesthetic & \\ \hline
Without content preference  & 65.80 & 56.83 & 54.67 & 94.67 & 65.73 & 73.93 & 68.61 \\
Without chain-of-speech     & 73.60 & 63.99 & \textbf{66.00} & 94.00 & 47.33 & \textbf{74.53} & 69.91 \\ \hline
Full system             & \textbf{74.82} & \textbf{68.38} & \textbf{66.00} & \textbf{96.67} & \textbf{66.40} & 73.60 & \textbf{74.31} \\ \hline
\end{tabular}

\caption{Ablation results on key model components. Results on a 30-sample subset of the PSP dataset.}  

    \label{tab:ablation}
\end{table*}

\subsection{Experimental Results}
\noindent \textbf{Quantitative results.} 
From Tab.~\ref{tab:main_experiment}, it can be observed that: (1) No method achieves an overall average score above 80\%, highlighting the inherent difficulty of the preference-guided paper-to-slides generation task. (2) Our method (GPT-4.1) achieves the highest overall score (75.8\%) and consistently outperforms all baselines across both preference-based and preference-independent metrics, suggesting that our framework produces slides that are not only well-aligned with user intent but also more informative and coherent from a general perspective. (3) Our approach also performs competitively when using the open-source Qwen2.5 + Qwen2.5VL models, demonstrating strong adaptability across different MLLM backbones without requiring model-specific tuning.

\noindent \textbf{Cost analysis.}
We sample five instances, each generating a 10-slide deck, with an average cost of \$0.665 (GPT version) or \$0.016 (Qwen version), based on OpenRouter~\cite{openrouter2025} API pricing as of October 13, 2025.

\noindent \textbf{Qualitative analysis.} 
Due to space limitations, detailed visualizations are included at our project website. Here we summarize key observations: 
AutoPresent~\cite{autopresent_cvpr25} fails to reflect aesthetic preferences due to its text-only input format. Although it can generate interleaved image-text output, the generated images are not faithfully derived from the paper, leading to weaker informativeness and potentially misleading content. ChatGPT supports multi-modal inputs but still struggles to consistently capture the desired visual style, and often omits figures and tables from the original paper, likely due to long context and the difficulty of extracting meaningful visuals from the paper. PPTAgent better preserves layout templates but still lacks alignment with the content structure. It also has a higher failure rate for image-related content extraction compared to our method. It frequently produces slides with large areas of unreasonable blank space and sometimes retains placeholder elements from the template that should have been removed. These observations highlight the challenge of slide generation and the effectiveness of our method.

\noindent \textbf{Human evaluation.} 
We recruited four graduate students with over two years of AI research experience to compare our method with PPTAgent, the strongest existing approach. Each participant completed 15 case studies. For each study, they were given the full input context (i.e., the target paper, the paper-slides sample pair, and the \texttt{.pptx} template), as well as anonymized outputs from both systems. They were asked to score each case on metrics mirroring those in our MLLM-based evaluation, covering two preference-based and two preference-independent metrics, as well as selecting an overall preferred output. The scoring rubric and instructions were identical to those used in the automatic evaluation. In total, we collected 60 independent human ratings, with each of the 30 unique cases evaluated by two evaluators. Our method achieved an 81.63\% win rate when compared to PPTAgent, demonstrating its superiority and consistent with evaluation by MLLMs. We also examined the agreement between human ratings and MLLM-based evaluations, observing an average Pearson correlation of 0.64 (with 0.5 generally considered a strong correlation). Further details are provided at our project website.

\subsection{Ablation Studies}

To assess the effectiveness of different components, we randomly sampled 30 cases and conducted ablations on two variants: (1) removing content preference guidance, and (2) disabling the chain-of-speech mechanism.

From Tab.~\ref{tab:ablation}, we highlight two key observations: (1) Removing content preference distillation notably degrades performance across all preference-based metrics, especially coverage, flow, and content structure (by around 10\%).
This not only validates the core hypothesis that modeling user-specific content preferences, even from implicit and unlabeled examples, is essential for generating slides aligned with communicative intent, but also demonstrates the effectiveness of our preference distillation module in capturing and leveraging such nuanced user signals. (2) Disabling the chain-of-speech module results in a clear drop in overall performance, especially in general content quality (66.4\% → 47.3\%). This underscores the importance of aligning slide planning with anticipated narration to improve clarity and informativeness. 

\section{Conclusion and Limitations}

In this paper, we explore the subjective nature of paper-to-slides generation. We propose a practical yet challenging task conditioned on user preferences captured through natural, real-world inputs. We introduce a human-like agentic framework that distills implicit preferences and progressively generates editable slides. A novel chain-of-speech mechanism bridges slide planning with oral narration, enhancing coherence and enabling downstream applications like video presentation. We also construct a benchmark dataset that simulates diverse user preferences and design interpretable metrics for robust evaluation. Experiments show the superiority of our method in both preference alignment and overall generation quality, paving the way for more personalized and flexible slide generation.

However, several limitations remain. First, our benchmark focuses exclusively on scientific papers. Extending it to broader domains (e.g., business reports, educational materials, advertising content) could benefit more fields. Second, while our training-free framework offers strong flexibility and adaptability, exploring end-to-end multimodal training for preference-guided slide generation is a promising direction. Third, although our MLLM-based evaluation shows general alignment with human judgment, a noticeable gap remains. We observe that MLLMs lack the fine-grained perception of humans and exhibit inherent self-bias, whereas cross-judge evaluations (e.g., Qwen judging GPT-based models) tend to yield results more consistent with human ratings. Designing more human-aligned evaluation protocols remains a valuable direction for future research.

\section*{Acknowledgments}
This research is supported by the National Research Foundation Singapore under its AI Singapore Programme (Award Number: AISG3-RP-2022-030).

\bibliography{aaai2026}

\clearpage
\UseRawInputEncoding
\appendix
\section*{Appendix}

We provide supplementary material for deeper understanding and analysis, arranged as follows:\\
1. Qualitative Results and Analysis (Sec.~\ref{sec:qualitative})\\
2. Details of Human Evaluation (Sec.~\ref{sec:human_eval})\\
3. Implementation Details (Sec.~\ref{sec:implementation})\\
4. Evaluation Metrics of the PSP Benchmark (Sec.~\ref{sec:eval_prompts})\\
5. Details of Baselines for Comparison (Sec.~\ref{app:baselines})\\
6. Downstream Applications (Sec.~\ref{sec:downstream_sup})\\

\begin{figure*}[htbp] 
    \centering
    \includegraphics[width=0.92\textwidth]{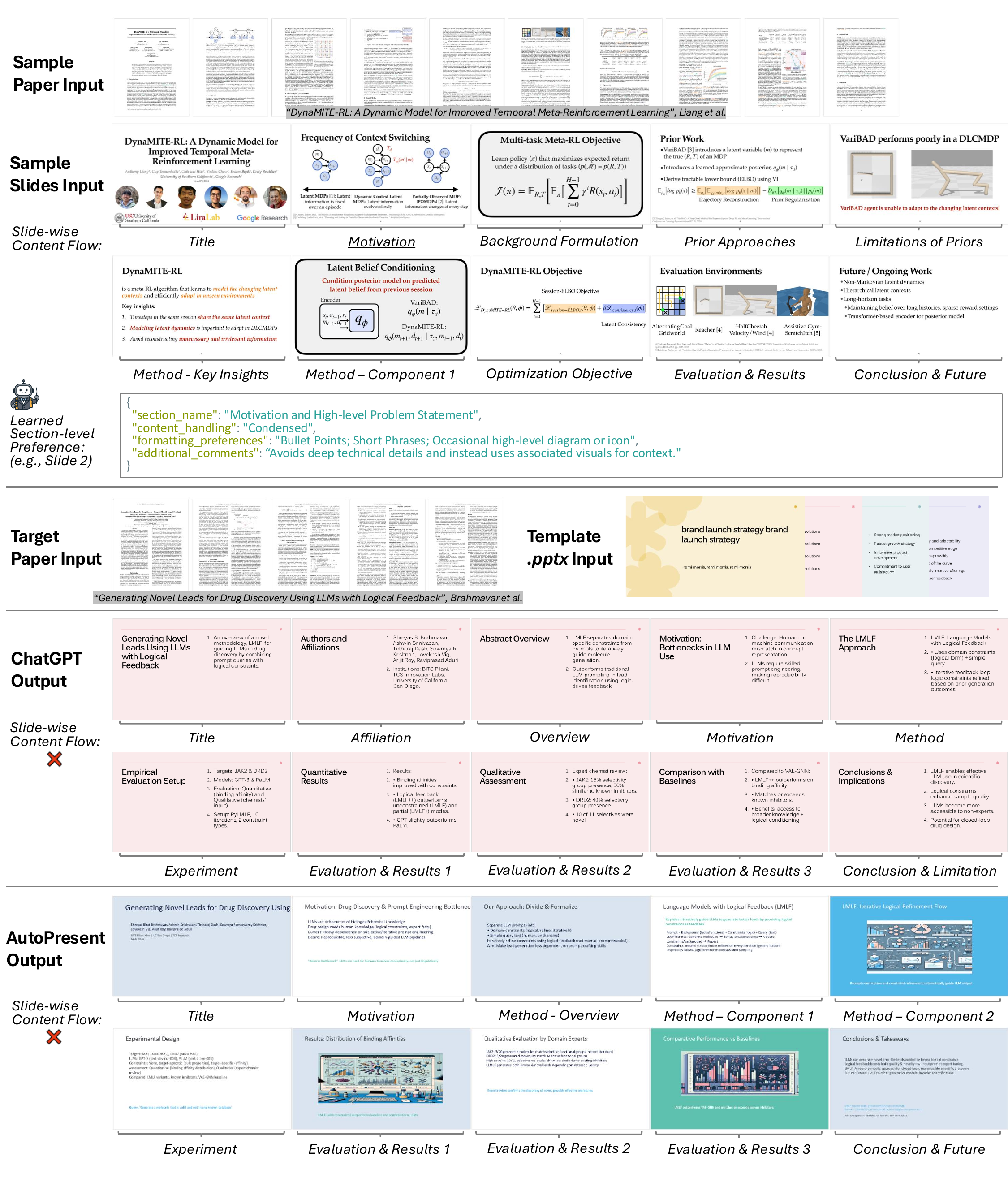}
    \caption{%
    Content structure analysis example with all system inputs.  
    \textbf{Top:} Sample paper \& slides, automatically extracted content structure flow, and detailed section-level preference learned from our SlideTailor.
    \textbf{Middle:} The input target paper and template \texttt{.pptx} file.
    \textbf{Bottom:} The output slides from baseline ChatGPT and AutoPresent, and corresponding content structure flow.}
    \label{fig:content_flow_analysis_1}
\end{figure*}

\begin{figure*}[htbp] 
    \centering
    \includegraphics[width=0.92\textwidth]{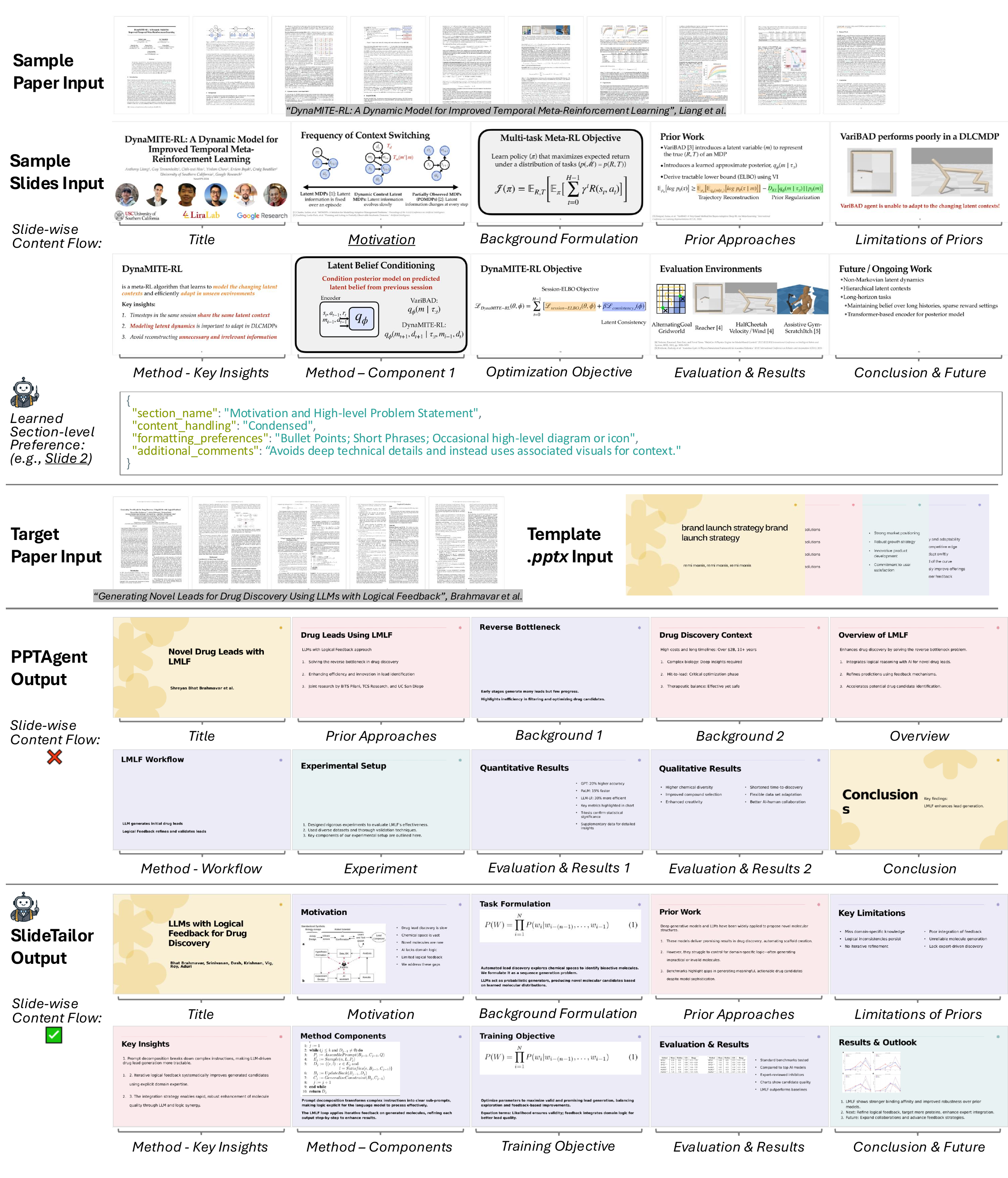}
    \caption{%
    Content structure analysis example with all system inputs.  
    \textbf{Top:} Sample paper \& slides, automatically extracted content structure flow, and detailed section-level preference learned from our SlideTailor.
    \textbf{Middle:} The input target paper and template \texttt{.pptx} file.
    \textbf{Bottom:} The output slides from baseline PPTAgent and our SlideTailor, and corresponding content structure flow. Together, these examples illustrate how SlideTailor preserves both high-level structural ordering and fine-grained content preferences when generating personalized slides.}
    \label{fig:content_flow_analysis_2}
\end{figure*}

\begin{figure*}[htbp]
    \centering
    \begin{subfigure}{\linewidth}
        \centering
        \includegraphics[width=\linewidth]{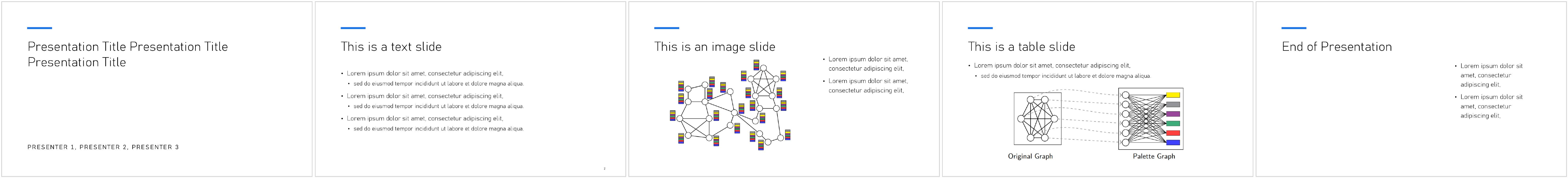}
        \caption{Template \texttt{.pptx} file input (truncated to 5 slides).}
    \end{subfigure}

    \begin{subfigure}{\linewidth}
        \centering
        \includegraphics[width=\linewidth]{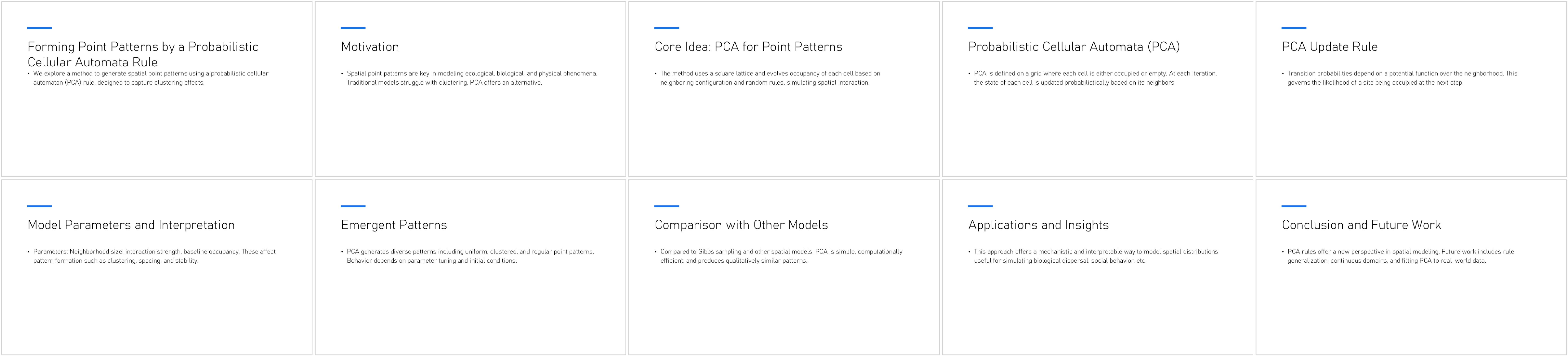}
        \caption{ChatGPT}
    \end{subfigure}
    
    \begin{subfigure}{\linewidth}
        \centering
        \includegraphics[width=\linewidth]{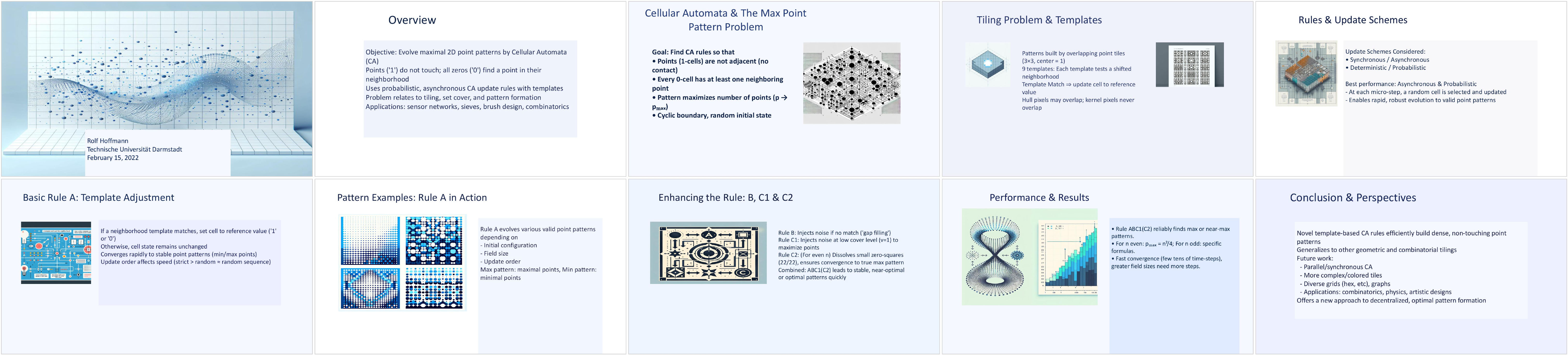}
        \caption{Autopresent}
    \end{subfigure}
    
    \begin{subfigure}{\linewidth}
        \centering
        \includegraphics[width=\linewidth]{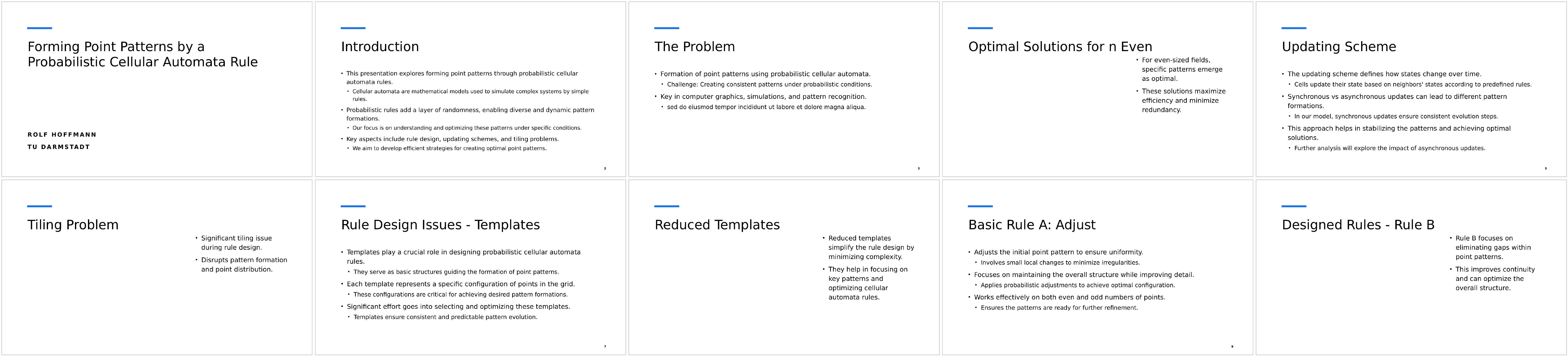}
        \caption{PPTAgent}
    \end{subfigure}
    
    \begin{subfigure}{\linewidth}
        \centering
        \includegraphics[width=\linewidth]{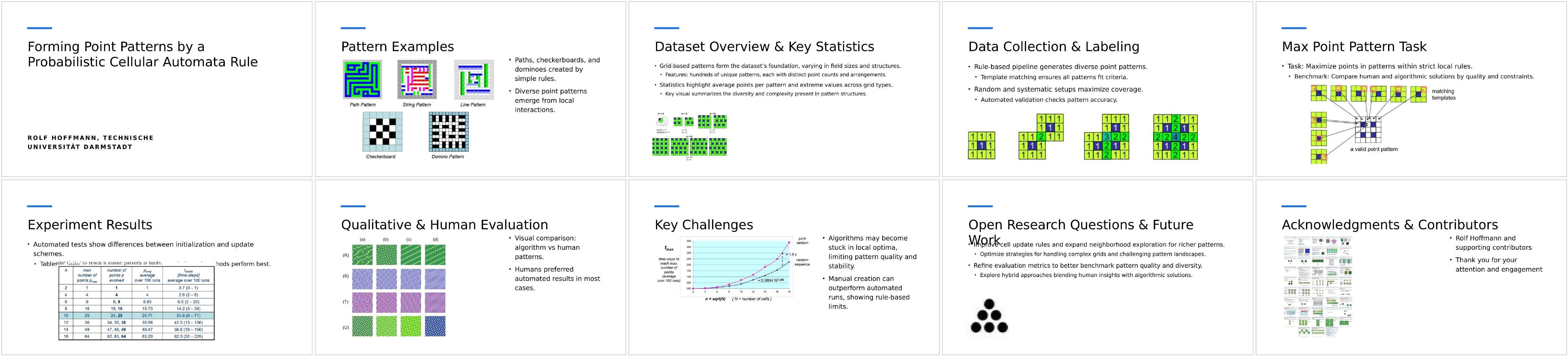}
        \caption{SlideTailor (Ours)}
    \end{subfigure}

    \caption{Examples of generated slides for qualitative analysis. Please zoom in to view the details.}
    \label{fig:qualitative_1}
\end{figure*}

\begin{figure*}[htbp]
    \centering

    \begin{subfigure}{\linewidth}
        \centering
        \includegraphics[width=\linewidth]{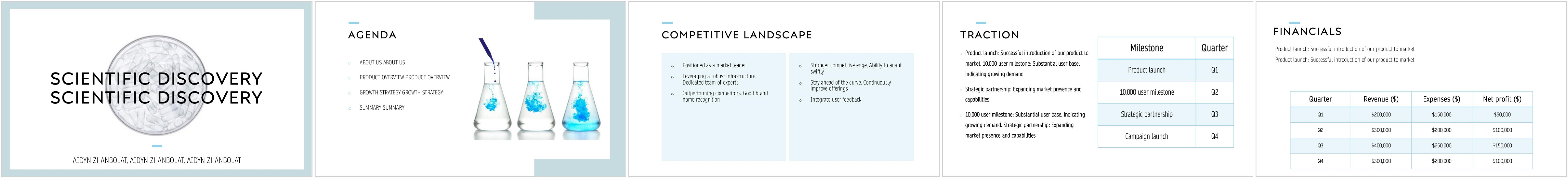}
        \caption{Template \texttt{.pptx} file input (truncated to 5 slides).}
    \end{subfigure}

    \begin{subfigure}{\linewidth}
        \centering
        \includegraphics[width=\linewidth]{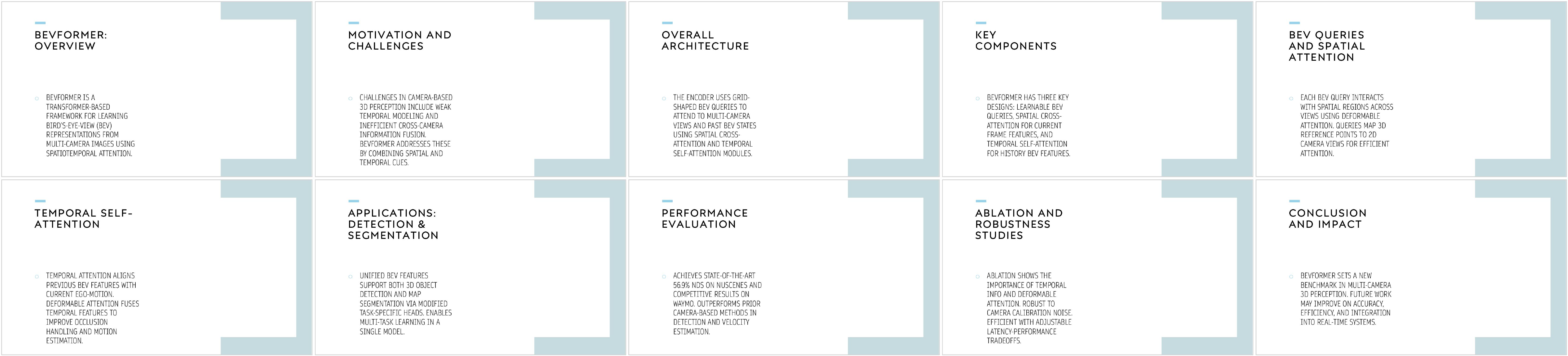}
        \caption{ChatGPT}
    \end{subfigure}
    
    \begin{subfigure}{\linewidth}
        \centering
        \includegraphics[width=\linewidth]{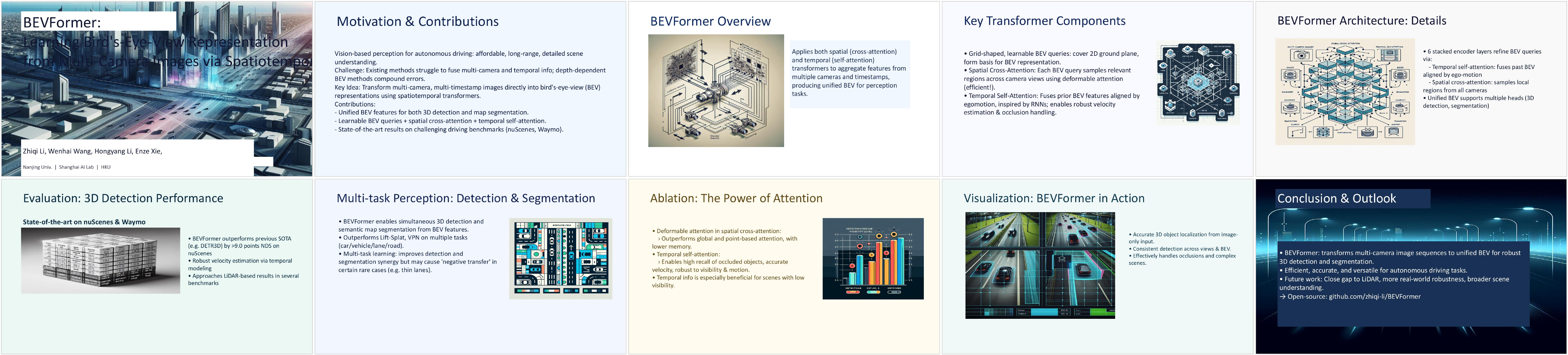}
        \caption{Autopresent}
    \end{subfigure}
    
    \begin{subfigure}{\linewidth}
        \centering
        \includegraphics[width=\linewidth]{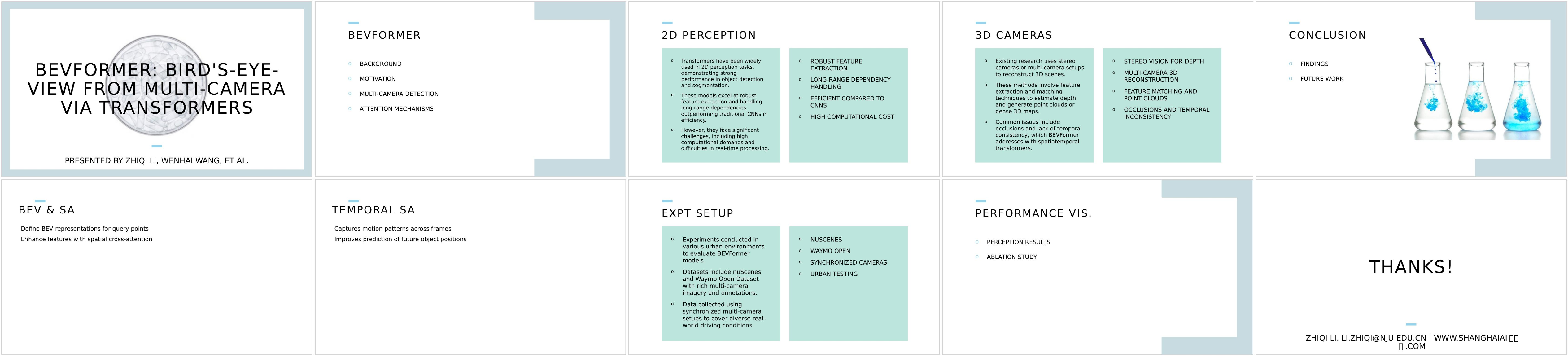}
        \caption{PPTAgent}
    \end{subfigure}
    
    \begin{subfigure}{\linewidth}
        \centering
        \includegraphics[width=\linewidth]{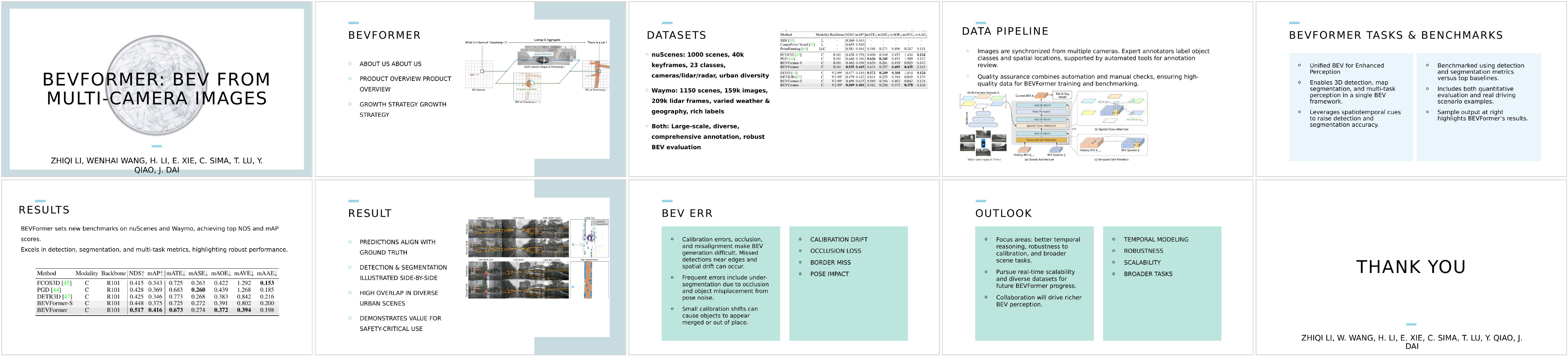}
        \caption{SlideTailor (Ours)}
    \end{subfigure}

    \caption{Examples of generated slides for qualitative analysis. Please zoom in to view the details.}
    \label{fig:qualitative_2}
\end{figure*}

\begin{figure*}[htbp]
    \centering
    \begin{subfigure}{\linewidth}
        \centering
        \includegraphics[width=\linewidth]{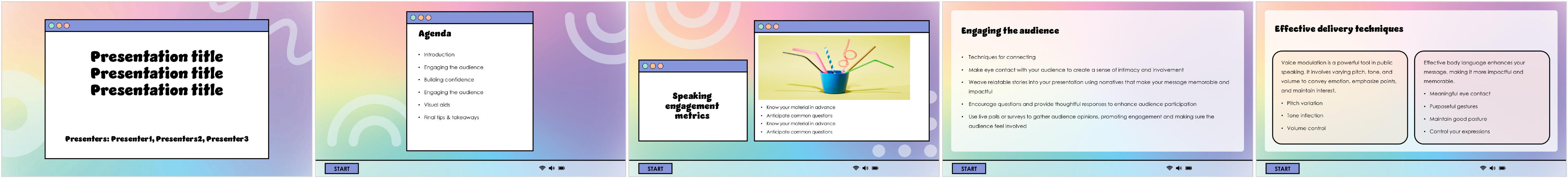}
        \caption{Template \texttt{.pptx} file input (truncated to 5 slides).}
    \end{subfigure}

    \begin{subfigure}{\linewidth}
        \centering
        \includegraphics[width=\linewidth]{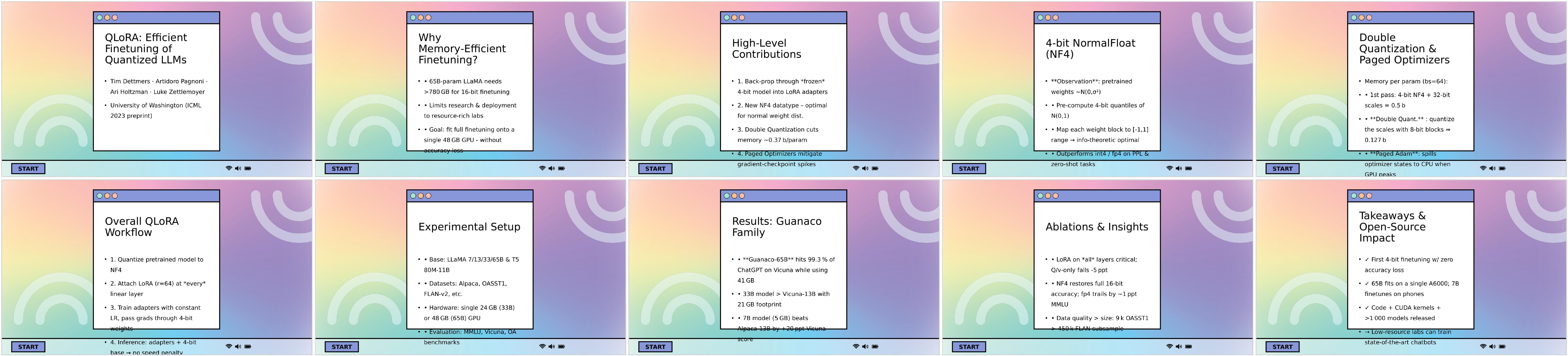}
        \caption{ChatGPT}
    \end{subfigure}
    
    \begin{subfigure}{\linewidth}
        \centering
        \includegraphics[width=\linewidth]{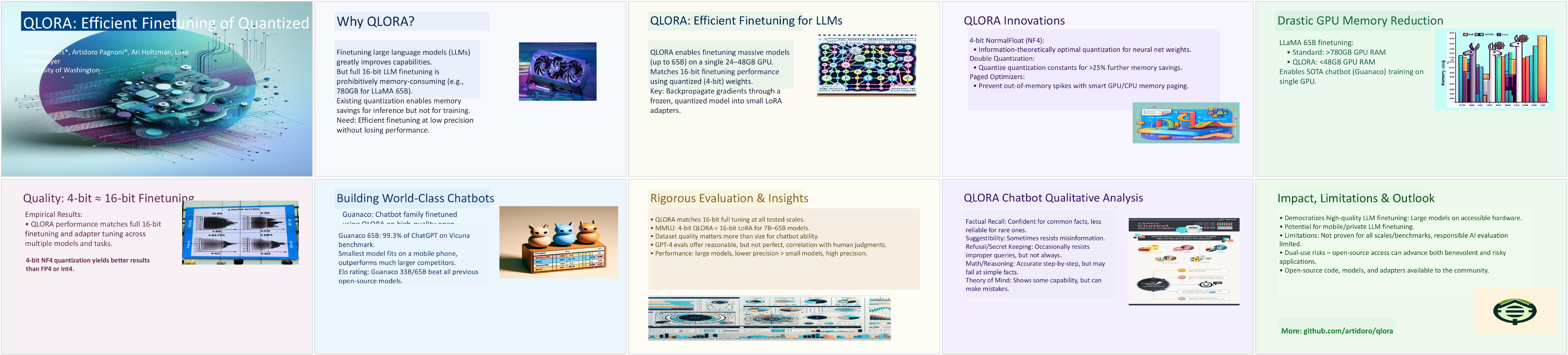}
        \caption{Autopresent}
    \end{subfigure}
    
    \begin{subfigure}{\linewidth}
        \centering
        \includegraphics[width=\linewidth]{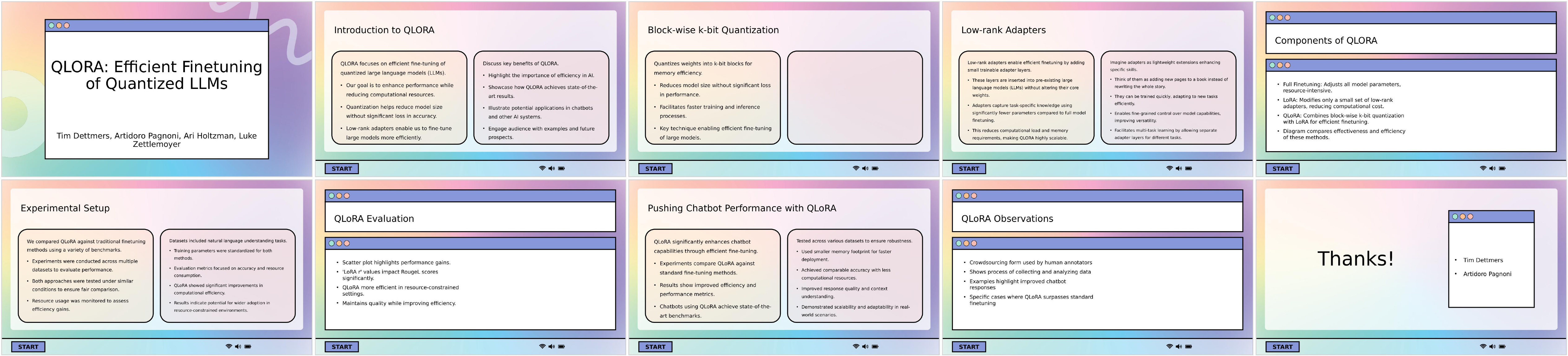}
        \caption{PPTAgent}
    \end{subfigure}
    
    \begin{subfigure}{\linewidth}
        \centering
        \includegraphics[width=\linewidth]{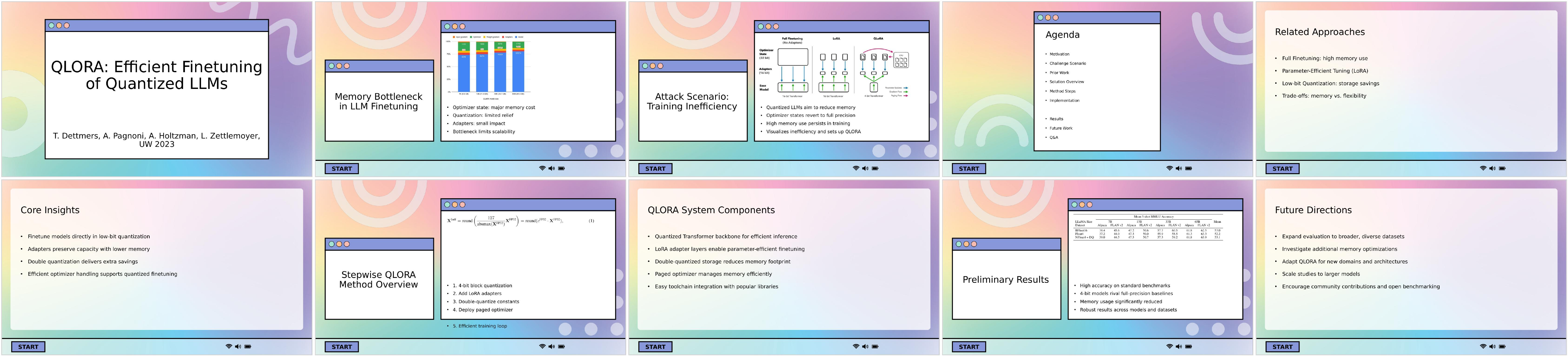}
        \caption{SlideTailor (Ours)}
    \end{subfigure}

    \caption{Examples of generated slides for qualitative analysis. Please zoom in to view the details.}
    \label{fig:qualitative_4}
\end{figure*}

\section{Qualitative Results and Analysis}\label{sec:qualitative}

Here we present detailed visualization and analysis of the generation results produced by our proposed SlideTailor and by existing methods.

Fig.~\ref{fig:content_flow_analysis_1} and \ref{fig:content_flow_analysis_2} show (i) the complete input (a sample paper–slides pair, a target paper, and a \texttt{.pptx} template) and (ii) the slide-wise presentation flows extracted from each method’s output slides. The slides generated by SlideTailor exhibit strong alignment with the sample paper-slides pair and template in both content structure and overall style.

\paragraph{Preference-independent analysis: general aesthetic and content quality.}
SlideTailor produces slides with noticeably higher visual quality and richer information. Rather than relying on text and images in isolation, it learns to combine textual and visual elements jointly to support clearer presentations. The resulting slides use a diverse yet harmonious color scheme inherited from the template and introduce subtle background shapes that enhance visual appeal. Moreover, the plots we include are extracted directly from the target paper and match the accompanying text, thus avoiding the misuse of generated graphics seen in AutoPresent~\cite{autopresent_cvpr25}.

\paragraph{Preference-based analysis: aesthetic alignment.}
Our generated slides adhere closely to the template’s background design, color palette, fonts, layout grids, and graphic motifs, with no significant deviation. Prior methods either under-utilize the diverse input template (ChatGPT, AutoPresent) or fail to learn style preferences from the sample paper–slides pair (AutoPresent, PPTAgent). In contrast, SlideTailor (e.g., Fig.~\ref{fig:qualitative_1}) mirrors the sample’s concise, example-led style: it introduces tables and charts at key points and reserves extended discussion for qualitative comparisons and challenges.

\paragraph{Preference-based analysis: content-structure alignment.}
Fig.~\ref{fig:content_flow_analysis_2} visualizes the slide-wise flow extracted from the sample slides, which follow a typical research-talk pattern: \textit{Title} $\rightarrow$ \textit{Motivation} $\rightarrow$ \textit{Background} $\rightarrow$ \textit{Prior Methods and Limits} $\rightarrow$ \textit{Method} - \textit{Key Insights} $\rightarrow$ \textit{Method – Components} $\rightarrow$ \textit{Method} – \textit{Optimization Objective} $\rightarrow$ \textit{Evaluation Results} $\rightarrow$ \textit{Conclusion and Future Work}. While the baseline PPTAgent preserves partial macro sections and re-orders \textit{Prior Approaches} and \textit{Background}, SlideTailor reproduces almost all sections in the correct order. This demonstrates that our global content-flow preference successfully guides high-level sequencing of the presentation.

Additional visual results in Fig. \ref{fig:qualitative_1}, \ref{fig:qualitative_2}, and \ref{fig:qualitative_4} compare slides generated by our method and by baseline methods alongside the input \texttt{.pptx} templates.

\begin{table}[t]
\setlength{\tabcolsep}{4pt}  
\centering
\scriptsize
\begin{tabular}{l|cc|cc|c} \hline
& \multicolumn{2}{|c|}{Preference-based} & \multicolumn{2}{c|}{Preference-independent} & \multirow{2}{*}{Average} \\ 
\multicolumn{1}{l|}{} & Content Struture     & Aesthetic     & Content             & Aesthetic            &                          \\ \hline
PPTAgent                & 2.07                 & 3.83          & 2.40                & 3.08                 & 2.85                     \\ \hline

Ours                & \textbf{3.10}                 & \textbf{3.95}          & \textbf{3.13}                & \textbf{3.48}                 & \textbf{3.42}                     \\ \hline         
\end{tabular}
\caption{Comparison of human scoring results (1-5 scale) across methods.}
\label{tab:pptagent_vs_ours_human}
\end{table}

\begin{table}[t]
\centering
\footnotesize
\begin{tabular}{lcccc} \hline
\multicolumn{1}{l}{} & Win & Lose & Tie & Win Rate \% (Tie Excluded) \\ \hline
PPTAgent             & 9   & 40   & 11  & 18.37                   \\
Ours          & 40  & 9    & 11  &     \textbf{81.63}              \\ \hline
\end{tabular}
\caption{Overall human preference comparison.}
\label{tab:win_rate}
\end{table}

\begin{table*}[t]
\centering
\footnotesize
\begin{tabular}{c|c|cc} \hline
Category                                & Metric           & Pearson r-value & Pearson p-value            \\ \hline
\multirow{2}{*}{Preference-based}  & Content Structure & 0.683           & $<$ 0.01 \\
       & Aesthetic        & 0.639           & $<$ 0.01 \\ \hline
\multirow{2}{*}{Preference-independent} & Content          & 0.602           & $<$ 0.01 \\
  & Aesthetic        & 0.626           & $<$ 0.01 \\ \hline
Average                                 & Average          & 0.638           & $<$ 0.01 \\ \hline
\end{tabular}
\caption{The Pearson correlation scores between human ratings and automatic MLLM ratings (GPT-4.1).}
\label{tab:human_llm_correlation}
\end{table*}

\begin{table*}[t]
\centering
\setlength{\tabcolsep}{5pt}  
\footnotesize
\begin{tabular}{ccccc} \hline
Category                                & Metric           & Avg. Absolute Difference & Var. Absolute Difference & \%Within 1-Point Difference \\ \hline
\multirow{2}{*}{Preference-based}       & Content Struture & 0.73                     & 0.53                     & 86.67                        \\
                                        & Aesthetic        & 0.95                     & 0.61                     & 78.33                        \\ \hline
\multirow{2}{*}{Preference-independent} & Content          & 0.80                     & 0.83                     & 83.33                        \\
                                        & Aesthetic        & 0.87                     & 0.82                     & 81.67                        \\ \hline
Average                                 & Average          & 0.84                     & 0.70                     & 82.50          \\ \hline             
\end{tabular}
\caption{Inter-annotator agreement.}
\label{tab:inter_human_agreement}
\end{table*}

\begin{table}[t]
\setlength{\tabcolsep}{4pt}  
\centering
\scriptsize
\begin{tabular}{c|cc|cc|c} \hline
& \multicolumn{2}{|c|}{Preference-based} & \multicolumn{2}{c|}{Preference-independent} & \multirow{2}{*}{Average} \\ 
\multicolumn{1}{l|}{} & Content Struture     & Aesthetic     & Content             & Aesthetic            &                          \\ \hline
Human                & 3.10                 & 3.95          & 3.13                & 3.48                 & 3.42                     \\
MLLM                  & 3.60                 & 4.83          & 3.39                & 3.80                 & 3.90         \\ \hline           
\end{tabular}
\caption{Comparison of human and MLLM-based (GPT-4.1) scoring (1-5 scale).}
\label{tab:human_vs_llm_eval}
\end{table}

\section{Details of Human Evaluation} \label{sec:human_eval}

\subsection{Detailed Setup}

We recruited four graduate students with over two years of AI research experience to compare our method against PPTAgent~\cite{pptagent}, the strongest prior approach in a model-agnostic manner. 
Each participant conducted 15 case studies, where they were given the full input context (i.e., the target paper, the paper-slides sample pair, and the \texttt{.pptx} template), as well as anonymized outputs from both systems for evaluation. 
Annotators were asked to score each case using selected metrics aligned with those in the MLLM-based evaluation, covering both preference-based (content structure, aesthetic) and preference-independent (content, aesthetic) aspects. The scoring instructions and rubric were kept consistent with those used in the automatic evaluation. 
In addition to scoring based on the specified metrics, we also asked annotators to indicate which of the two anonymized systems they preferred overall and briefly explained their reasoning. This step is intended to encourage human evaluators to reflect more carefully on their scoring decisions.
To reduce inter-annotator subjectivity, each case was evaluated by two different annotators. In total, we collected 60 case-level responses covering 30 distinct cases.

\subsection{Detailed Scoring Results and Analysis}
We compare human ratings of SlideTailor and the strong counterpart PPTAgent across all cases and metrics. As shown in Tab.~\ref{tab:pptagent_vs_ours_human}, SlideTailor consistently outperforms PPTAgent. Besides, the overall preference provided by humans (Tab.~\ref{tab:win_rate}) shows that in 81.63\% of cases, the slides generated by our SlideTailor are preferred. These results further validate the superiority of our method.

For the correlation between MLLM evaluation and human scoring,
as shown in Tab.~\ref{tab:human_llm_correlation}, the Pearson r values exceed 0.6 across different metrics, indicating a strong correlation (as values above 0.5 are generally considered substantial). This demonstrates the validity of our designed automatic evaluation metrics.

Despite its effectiveness, we also observe limitations of MLLM-based evaluation when applied to the preference-guided paper-to-slides generation task. Specifically, we found that human evaluators tend to assign lower scores compared to the automatic ratings from MLLM models, as shown in Tab.~\ref{tab:human_vs_llm_eval} (evaluated on the same subset of results generated by our SlideTailor), indicating that human judges are generally more stringent. We attribute this to the current limitations of MLLMs in capturing fine-grained visual details, such as font styles, subtle layout issues, and precise text recognition. This gap is particularly evident in aesthetic-related metrics, where problems like inappropriate font choices or overlapping elements may go unnoticed by the model but are easily flagged by human evaluators. Therefore, we believe that enhancing the fine-grained perceptual ability of MLLMs for slide evaluation represents a promising direction for future research. Besides, we also observe that cross-judge evaluations (e.g., Qwen judging GPT-based models) tend to yield results more aligned with human ratings, which also serves as a potential measure to reduce self-bias in MLLM evaluation.

\subsection{Inter-Annotator Agreement}
As shown in Tab.~\ref{tab:inter_human_agreement}, due to the inherent subjectivity of human judgment, different annotators may assign different scores to the same case. However, the differences remain within an acceptable range: the average absolute difference is below 1, and in 82.5\% of the cases, the score difference is less than 1 point on a 1-5 scale.

\section{Implementation Details} \label{sec:implementation}

\subsection{Implicit Preference Distillation} \label{app:pref_dist}
\noindent \textbf{Content preference.}
Here we provide the specific prompt used to instruct the LLM to infer the latent mapping function, as illustrated in Fig.~\ref{fig:prompt_implicit_preference}.

\noindent \textbf{Aesthetic preference.}
Here we provide the instruction used for aesthetic preference distillation, as shown in Fig.~\ref{fig:prompt_layout_preference_distillation}.

\subsection{Preference-Guided Slide Planning}
\noindent \textbf{Conditional paper reorganization.} The instruction for paper reorganization is shown in Fig.~\ref{fig:prompt_document_reorganizer}.

\noindent \textbf{Slide-wise outline generation with chain-of-speech.} The instruction for content outline generation is shown in Fig.~\ref{fig:prompt_content_outline}.
 
\noindent \textbf{Template-aware layout selection.} The instruction for layout selection is shown in Fig.~\ref{fig:prompt_layout_selection}.

\section{Evaluation Metrics of the PSP Benchmark}
\label{sec:eval_prompts}

\subsection{Preference-based Rubrics}

Fig. \ref{fig:eval_pref_content_structure_prompt} and \ref{fig:eval_pref_aesthetic_prompt} present the prompts for evaluating content structure and aesthetic similarity in the preference-based MLLM-as-a-judge evaluation.

\subsection{Preference-independent Rubrics}

Fig. \ref{fig:eval_content_structure_prompt} and \ref{fig:eval_aesthetic_prompt} show the prompts for content and aesthetic evaluation in the preference-independent MLLM-as-a-judge setting.

\section{Details of Baselines for Comparison} \label{app:baselines}
\subsection{Prompt for ChatGPT}
For the ChatGPT baseline, we used GPT-4o via the OpenAI ChatGPT web interface\footnote{\url{https://chatgpt.com/}}. The target paper, sample paper-slides pair, and template \texttt{.pptx} are uploaded with the interface’s file-attachment feature. We also supply a slide generation prompt to simulate our conditional generation process. The complete prompt is shown in Fig. \ref{fig:chatgpt_prompt}.

\subsection{Prompt for AutoPresent}
We instantiate AutoPresent with GPT-4.1 as the LLM backend. Since it only accepts plain-text input, we extract complete raw text from the target paper, sample paper, and its slides, and inserted these strings into the prompt. The exact template used in our experiment is shown in Fig. \ref{fig:autopresent_prompt}.

\subsection{PPTAgent}
We use the official open-source implementation.

\section{Downstream Applications} \label{sec:downstream_sup}
\subsection{Conditional vs. Unconditional Generation}
The proposed preference-guided paper-to-slides generation task can be viewed as a form of conditional generation. Beyond producing slides that align with specific user preferences, our setting can naturally turn into a preference-independent (i.e., unconditional) generation scenario. For instance, when the input paper-slides pair and template represent general presentation convention, such as using a standard content structure and a plain whiteboard-style template, the resulting slides can be regarded as common, general-purpose slides. Therefore, when the additional conditional inputs are set to some default values, our SlideTailor can also effectively function in an unconditional generation mode. Such a setting further simplifies the user input requirements, offering greater flexibility.

\subsection{Video Presentation}

Beyond customized slide generation, we take a preliminary step towards speaker-aware video presentation by composing slides with synthesized narration and talking head videos. Given the generated slide deck and speech script from SlideTailor, we first convert the script into speech audio using a voice cloning method\footnote{\url{https://github.com/boson-ai/higgs-audio}}, which replicates the user's vocal identity from a short sample audio clip. Then, an audio-driven talking head model~\cite{memo} generates lip-synced facial animation from the synthesized audio, conditioned on a provided identity image of the user.
We overlay the talking head onto each slide (e.g., bottom-right corner) to produce slide-specific video segments, ensuring audiovisual synchronization. These segments are then concatenated in slide order, with transition such as pauses or fade effects to ensure smooth playback. All these composition steps, including overlaying, synchronizing, and applying transition, are implemented using FFmpeg. The final result is a fully synchronized, speaker-aware video presentation.

\begin{figure*}[htbp]
  \centering
\begin{lstlisting}[
  basicstyle=\fontsize{6}{5.3}\ttfamily,
  frame=single,        % ← draws one solid-line bounding box
  framerule=0.35pt,     %   thickness of the rule
  rulecolor=\color{black}, %   rule colour (default is black; can omit)
  framesep=6pt,        %   padding between text and the frame
  breaklines=true,        % automatic wrap
  breakatwhitespace=true, % try to break only at spaces
  breakautoindent=false,
  breakindent=0pt,
  columns=fullflexible,   % optional but avoids odd gaps
  inputencoding=utf8,         % tell listings the file is UTF-8
  literate={–}{{\textendash} }1
           {"}{{\textquotedblleft}}1
           {"}{{\textquotedblright}}1
           {’}{{\textquoteright}}1 % right single quote
           {‘}{{\textquoteleft}}1, % left single quote
]
You are a document transformation and summarization specialist, tasked with analyzing how a user transforms a research paper (PDF version) into a slides presentation.

You are given two reference files:
1. <Research Paper PDF>: This is the original research paper PDF.
2. <Corresponding Slides>: The corresponding slides presentation derived from the paper.

**Task Objective**:
Your goal is to extract a generalized presentation preference guideline based on how the research paper was transformed into the slides.

This preference guideline will later support conditional document summarization and slides generation, so focus on:
1. How content is selected, organized, expanded, condensed, or omitted;
2. How each section is formatted;
3. Any stylistic patterns specific to each section.

You may freely summarize general academic section topics commonly found in research papers (e.g., Task Introduction, Challenge, Motivation, Method, Experiments, Future Work, Conclusion),
but must not copy or quote any specific sentences, unique technical terms, method names, or dataset names from the research paper.

Special Notes:
You must produce high-level, generalizable descriptions.
Do not include specific method names, model names, dataset names, algorithm names, or paper-specific experimental setups in any part of the extracted structure (you can indicate those in the Additional Comments).
Section names and flow must be domain-agnostic and task-transferable.

**Specific Analysis Dimensions**:
1. Narrative Flow Preference:
  (1) List the logical sequence of major sections/topics as presented in the slides.
  (2) Only include sections that actually appear in the slides.

2. Section-Level Preferences (following the order of the narrative flow):
  (1) Content Handling
  Indicate how the content is treated in the slides compared to the research paper:
  a. Expanded: Elaborated more than in the research paper.
  b. Newly Added: Newly introduced content not explicitly existing in the research paper.
  c. Condensed: Summarized or simplified compared to the research paper.

  (3) Formatting Preferences
  Describe the formatting style adopted for this section:
  a. Bullet Points
  b. Short Phrases
  c. Full Sentences
  d. Visualization-heavy
  e. Minimal Text
  f. Other recognizable formatting patterns

  (4) Additional Comments
  Capture any stylistic features or special habits specific to this section, such as:
  a. Custom naming (e.g., using a method's name instead of the generic title "Method").
  b. Use of specific visual elements (e.g., heavy use of flowcharts, block diagrams, equations).
  c. Particularly detailed writing or highly concise expression.
  d. Other noticeable stylistic preferences (e.g., frequent highlighting, color usage, consistent framing style).
  e. Some typical content style that you think is necessary to capture, like what aspect is focused on as important, or what is the most important part of the section, details (e.g., hardware, dataset, setup, etc.), and experiments.

**Omitted Sections**:
List any major sections from the research paper that were completely omitted (i.e., no content or trace found in the slides).

**Output Format**:
{
  "presentation_guidelines": {
    "narrative_flow_preference": [
      "Ordered list of sections reflecting the logical storyline in the user's slides"
    ],
    "section_level_preferences": [
      {"section_name": "Title",
        "content_handling": "Condensed",
        "formatting_preferences": "Short Phrases / Full Sentences / Minimal Text / etc.",
        "additional_comments": "e.g. Full paper title with author name, affiliated institution, conference date and name."
      },
      {"section_name": "Task Introduction",
        "content_handling": "Expanded / Newly Added / Condensed",
        "formatting_preferences": "Bullet Points / Short Phrases / Full Sentences / Visualization-heavy / Minimal Text / etc.",
        "additional_comments": "e.g., Very detailed description style; uses custom title naming like 'Problem Overview', the task name."
      },
      {"section_name": "Challenge",
        "content_handling": "Expanded / Newly Added / Condensed",
        "formatting_preferences": "Bullet Points / Short Phrases / Full Sentences / Visualization-heavy / Minimal Text / etc.",
        "additional_comments": "e.g., Generalized stylistic observations (e.g., uses custom abstract titles, prefers flowcharts to text summaries, etc."
      },
      {"section_name": "Motivation",
        "content_handling": "Expanded / Newly Added / Condensed",
        "formatting_preferences": "Bullet Points / Short Phrases / Full Sentences / Visualization-heavy / Minimal Text / etc.",
        "additional_comments": "e.g., Summarized quickly; prefers very brief textual description without figures."
      },
      {"section_name": "Method",
        "content_handling": "Expanded / Newly Added / Condensed",
        "formatting_preferences": "Bullet Points / Short Phrases / Full Sentences / Visualization-heavy / Minimal Text / etc.",
        "additional_comments": "e.g., Summarized quickly; prefers very brief textual description without figures."
      },
      {"section_name": "Implementation Details",
        "content_handling": "Expanded / Newly Added / Condensed",
        "formatting_preferences": "Bullet Points / Short Phrases / Full Sentences / Visualization-heavy / Minimal Text / etc.",
        "additional_comments": "e.g., typical aspects, such as training setup, GPU, dataset, evaluation metrics, etc."
      }
      // Continue for other sections according to narrative flow
    ],
    "omitted_sections": [
      "Section names from the research paper that were completely omitted in the slides (e.g., Related Work, Detailed Implementation)"
    ]
  }
}

Input:
<Research Paper PDF Begins>:
{{reference content pdf}}
<Research Paper PDF Ends>.

<Corresponding Slides Begins>:
{{reference content slide}}
<Corresponding Slides Ends>.
\end{lstlisting}
  \caption{Instruction for latent mapping function inference during content preference distillation.}
  \label{fig:prompt_implicit_preference}
\end{figure*}

\begin{figure*}[htbp]
  \centering
\begin{lstlisting}[
  literate={–}{{\textendash} }1
           {"}{{\textquotedblleft}}1
           {"}{{\textquotedblright}}1
           {’}{{\textquoteright}}1 % right single quote
           {‘}{{\textquoteleft}}1 % left single quote
           {!}{{!}}1,
]
You are given the structured content of slides, where each slide is represented as a dictionary (e.g., "slide_0", "slide_1", ...) and each slide contains multiple elements (e.g., "shape_0", "shape_1", ...). Each element provides information such as description, size, position, and text content.

Your task:
For each slide, analyze all its elements and summarize the main theme of the slide in one concise sentence. The main theme should capture the core media type and purpose of the slide, and briefly describe the layout or content type, without mentioning technical or formatting details.

Input Example:
{
  "slide_0": {
    "shape_0": {
      "pptc_description": "[TextBox id=0]\n",
      "pptc_size_info": "Size: height=96pt, width=347pt\n",
      "pptc_space_info": "Visual Positions: left=47pt, top=324pt\n",
      "pptc_text_info": "Presenter: Presenter 1, Presenter 2, ..."
    }
  },
  "slide_1": {
    "shape_0": {
      "pptc_description": "[TextBox id=0]\n",
      "pptc_text_info": "Title Title Title Title\n"
    }
  }
}

Output Example:
{
  "slide_0": "Opening, introduce main title, author information, xxx",
  "slide_1": "Contents, pure text with short paragraphs",
  "slide_2": "Contents, pure text with multiple bullet points",
  "slide_3": "Contents, image with text, layout xxx",
  "slide_4": "Contents, equation with text, layout xxx",
  "slide_5": "Contents, table with text, like for Experimention, layout"
  ...
}

Requirements:
- For each slide, only output the main theme as a single sentence.
- Do not mention technical details.
- Focus on the actual content and its purpose.
- If the slide contains a mix (e.g., image and text), briefly mention both.
- Strictly follow the name of the keys (e.g., "slide_0", "slide_1", ...)

Input Format
<Structured Slide Info Begins>:
{{slide_info}}
<Structured Slide Info Ends>.

Output:
Only return the **structured JSON**.

\end{lstlisting}
    \caption{Instruction for aesthetic preference distillation.}
  \label{fig:prompt_layout_preference_distillation}
\end{figure*}

\begin{figure*}[htbp]
  \centering
\begin{lstlisting}[
  basicstyle=\fontsize{5.6}{4.6}\ttfamily,
  frame=single,        % ← draws one solid-line bounding box
  framerule=0.35pt,     %   thickness of the rule
  rulecolor=\color{black}, %   rule colour (default is black; can omit)
  framesep=3pt,        %   padding between text and the frame
  breaklines=true,        % automatic wrap
  breakatwhitespace=true, % try to break only at spaces
  breakautoindent=false,
  breakindent=0pt,
  columns=fullflexible,   % optional but avoids odd gaps
  inputencoding=utf8,         % tell listings the file is UTF-8
  literate={–}{{\textendash} }1
           {"}{{\textquotedblleft}}1
           {"}{{\textquotedblright}}1
           {’}{{\textquoteright}}1 % right single quote
           {‘}{{\textquoteleft}}1 % left single quote
           {!}{{!}}1,
]
## Role
You are a **document content divider and summarization specialist**, tasked with reorganizing research papers into a structured two-level JSON format based on user-specific presentation preferences.  
The output will serve as the direct foundation for later slide generation and speech draft generation.

## Inputs
- **<User Preference Guidelines>**  
  A JSON object generated from prior analysis of reference papers and slides. It includes:
  - Preferred narrative flow (general section ordering)
  - Section-level content handling instructions (Expanded / Newly Added / Condensed)
  - Formatting preferences for each section
  - Additional stylistic comments (e.g., title naming patterns, visual usage preferences)
- **<Target Paper>**  
  A new research paper to be summarized and reorganized according to the above user preferences.

## Task Objective
**Conditionally summarize and restructure** the <Target Paper> into a two-level JSON structure,  
**strictly aligned** with the <User Preference Guidelines>, while ensuring:
- Conciseness, informativeness, and clarity
- Logical narrative flow
- Stylistic consistency with user’s preference
- Faithfulness to the content of the Target Paper
- Readiness for direct slide generation

## Detailed Steps
### Step 1: Analyze User Preferences
- Carefully read and extract key aspects from the <User Preference Guidelines>:
  - Preferred narrative flow
  - Content handling tendencies
  - Formatting and stylistic conventions

### Step 2: Summarize and Reorganize the Target Paper
Following the user preferences:
- **(A) Detect Logical Content Blocks**  
  Identify logical sections/subsections based on the thematic structure of the Target Paper.
- **(B) Align to the Narrative Flow Preference**  
  Reorganize sections according to the preferred flow.  
  Minor adjustments are allowed for coherence, but the structure must reflect user intent.
- **(C) Apply Section Handling Instructions**  
  Expand, Condense, or Newly Add sections as guided by user preferences.
- **(D) Supplement Missing Sections If Needed**  
  - If sections important to the user's preference (e.g., Background, Task Setup) are absent, infer and create them logically based on Target Paper content.
  - Ensure any supplemented content remains faithful (must be relevant) to the actual context of the Target Paper.
- **(E) Refine Titles and Content**  
  - Titles: Clean, generalizable, intuitive; consistent with user naming tendencies.
  - Content: Rephrased and summarized appropriately.
- **(F) Prune Unnecessary Content**  
  - Also learn from the user preference guidelines to prune unnecessary content. But be careful, do not prune the content that is important for the slides and speech draft -> informative and longer is usually better.

## Output Format
Generate a structured JSON output with the following format:
Example Output:
{
    "metadata": {
        "title": "title of document",
        "author": "name of authors",
        "publish date": "date of publication",
        "organization": "name of organization"
    },
    "sections": [
        {
            "title": "title of section1",
            "subsections": [
                {
                    "title": "title of subsection1.1",
                    "content": "content of subsection1.1"
                },
                {
                    "title": "title of subsection1.2",
                    "content": "content of subsection1.2"
                }
            ]
        },
        {
            "title": "title of section2",
            "subsections": [
                {
                    "title": "title of subsection2.1",
                    "content": "content of subsection2.1"
                }
            ]
        }
    ]
}

## Execution Rules
- Prioritize the narrative flow specified in the <User Preference Guidelines>.
- Allow minor adjustments that can better align with both the current paper and preference.
- Supplement missing sections based on logical inference if important sections (e.g., Background, Task Setup) are absent but relevant in the Target Paper.
- Each time you generate a section or its content, you must check whether this content actually exists in the provided target paper, rather than relying solely on user preferences. User preferences should only inform the style, not introduce unrelated or extraneous content. This ensures that no irrelevant information is included.
- Content Generation Principle:
  - The content must be faithfully based on the Target Paper.
  - Do not introduce details from the Reference Paper that are irrelevant to the Target Paper's topic.
  - The User Preference only guides structure, content handling, and formatting style - it must not override the factual basis of the Target Paper.
- Title Refinement:
  - Can be refined according to the target paper, with the similar spirit of the user preference guidelines.
- Content Length:
  - Each subsection can be long to ensure informativeness, as this serves as the primary and only source for subsequent speech draft and slides generation. In other words, we prioritize comprehensiveness and detail at this stage, and you can only filter out content that you believe will definitely not be used in either the slides or speech draft. This approach allows for a more thorough initial capture of information while maintaining relevance to the presentation goals.
- Final Output Requirement:
  - Output only the final structured JSON, without any additional commentary, notes, or explanation outside the JSON.

Input Format
<User Preference Guidelines Begins>:
{{user preference guidelines}}
<User Preference Guidelines Ends>.

<Target Paper Begins>:
{{target paper}}
<Target Paper Ends>.

Your goal is to analyze how the user transformed the <Reference Content PDF> into the <Reference Content Slide>, extract key user preferences, and apply them to summarize and restructure the <Target Paper> into a informative and structured JSON format for later PPT generation.

Output:
Only return the **structured JSON** of the **Target Paper** as per the format above, **applying** the guidelines **to structure, flow, condense, and prioritize** the content.
\end{lstlisting}
    \caption{Instruction for paper reorganization.}
    \label{fig:prompt_document_reorganizer}
\end{figure*}

\begin{figure*}[htbp]
  \centering
\begin{lstlisting}[
  basicstyle=\fontsize{6.4}{5.2}\ttfamily,
  frame=single,        % ← draws one solid-line bounding box
  framerule=0.35pt,     %   thickness of the rule
  rulecolor=\color{black}, %   rule colour (default is black; can omit)
  framesep=6pt,        %   padding between text and the frame
  breaklines=true,        % automatic wrap
  breakatwhitespace=true, % try to break only at spaces
  breakautoindent=false,
  breakindent=0pt,
  columns=fullflexible,   % optional but avoids odd gaps
  inputencoding=utf8,         % tell listings the file is UTF-8
  literate={–}{{\textendash} }1
           {"}{{\textquotedblleft}}1
           {"}{{\textquotedblright}}1
           {’}{{\textquoteright}}1 % right single quote
           {‘}{{\textquoteleft}}1 % left single quote
           {!}{{!}}1,
]
You are a professional presentation content designer. Your task is to generate a structured presentation outline in JSON format, focusing primarily on content coherence and quality. All content must be based on the summarized document and should reflect any user preferences if provided. 

Instructions:
    1. Review the provided document overview, image captions, and user preference guidelines.
    2. All slide content (including topic, content, and speech draft) must be based on the summarized document, and should also consider user preferences if available.
    3. Focus on content coherence, logical flow, and information organization across slides.
    4. Ensure content is comprehensive yet concise according to user preferences.
    5. Create compelling speech drafts that effectively communicate the key points.
    6. Strictly follow the user preference guidelines (if provided) to make the presentation content more tailored.
For each slide, provide the following fields:
    - Slide Topic: A high-level summary of the slide's objective or topic, used as the key in the JSON, and also the title of the slide!
    - purpose: A concise explanation of the communicative goal or function of the slide - not just what is shown, but why this slide exists.
    - speech_draft: A simulated speech draft for this slide.
    - subsections: Relevant subsection titles related to the slide's content.
    - image_assets: Path(s) to the image that best supports the content.
    - content_style: The preferred style or level of detail for the text displayed on this slide (e.g., concise bullet points, detailed explanations, or conversational tone), as learned from the given user preferences. This refers specifically to the text that will actually appear on the slide, which may differ in granularity or style from the speech draft, as they serve different purposes.
    - layout_recommendation: A brief recommendation of what type of layout would best suit this content (e.g., "text-heavy", "image with text", "comparison", "list", "title only").
    
Please ensure that the content of all fields-including Slide Topic, purpose, speech draft, and Subsections-is consistent and coherent with each other. The information presented in each field should logically align and reinforce the overall message of the slide, maintaining a unified and seamless narrative throughout the entire presentation.

If you determine that the final slide serves as a summary or contains substantive content-rather than simply signaling the end-this should be reflected in both the user preferences and the speech draft. In such cases, your layout_recommendation should suggest a content-oriented layout for the last slide, rather than a purely ending layout, to better accommodate the inclusion of meaningful information.

Note: 
    - The preferred conciseness of the text displayed on the slide, as indicated by user preferences, does not necessarily apply to the speech draft. The speech draft should remain as coherent, clear, and informative as possible, providing sufficient detail for understanding. Additionally, both the slide content and the speech draft should be informative and meaningful, rather than containing unhelpful or superficial information.
    - Avoid using the same image for multiple times.
    - The order of the sections should remain consistent with both the provided user preferences and the summarized document. The title of each slide can be taken from the summarized document, with appropriate grouping or splitting as needed to meet the required number of slides.
    Output requirements:
    - Output must be in JSON format.
    - The number of slides in your output must exactly match: {{ num_slides }}
    - Focus on content quality and coherence at this stage, not specific layout selection.

Example output (for a {{ num_slides }}-slide presentation):
{
    "1_Opening of XX": {
        "purpose": "...",
        "speech_draft": "...",
        "subsections": [],
        "content_style": "...",
        "layout_recommendation": "title with subtitle"
    },
    "2_Introduction": {
        "purpose": "...",
        "speech_draft": "...",
        "subsections": ["Title of Subsection 1.1", "Title of Subsection 1.2"],
        "content_style": "...",
        "image_assets": ["path_to_the_image1"],
        "layout_recommendation": "text with bullets"
    },
    "3_XX Method": {
        "purpose": "...",
        "speech_draft": "...",
        "subsections": ["Title of Subsection 2.1", "Title of Subsection 2.2, ..."],
        "content_style": "...",
        "image_assets": ["path_to_the_image1", "path_to_the_image2"],
        "layout_recommendation": "text with supporting image"
    },
    "4_Visual example of XX": {
        "purpose": "...",
        "speech_draft": "...",
        "subsections": ["Title of Subsection 3.1"],
        "content_style": "...",
        "image_assets": ["path_to_the_image3", "path_to_the_image4", "..."],
        "layout_recommendation": "image-focused with caption"
    },
    ...,
    "{{num_slides}}_Ending of XX": {
        "purpose": "...",
        "speech_draft": "...",
        "subsections": [],
        "content_style": "...",
        "layout_recommendation": "summary with key takeaways"
    }
}

Input:
<Summarized Document Begins>
{{ summarized_doc_content }}
<Summarized Document Ends>

<User Preference Guidelines Begins>
{{ pref_guidelines }}
<User Preference Guidelines Ends>

<Available Images Begins>
{{ image_information }}
<Available Images Ends>

The desired number of slides: {{ num_slides }}

Your Output:
\end{lstlisting}
    \caption{Instruction for content outline generation.}
    \label{fig:prompt_content_outline}
\end{figure*}

\begin{figure*}[htbp]
  \centering
\begin{lstlisting}[
  basicstyle=\fontsize{7.2}{5.4}\ttfamily,
  frame=single,        % ← draws one solid-line bounding box
  framerule=0.35pt,     %   thickness of the rule
  rulecolor=\color{black}, %   rule colour (default is black; can omit)
  framesep=6pt,        %   padding between text and the frame
  breaklines=true,        % automatic wrap
  breakatwhitespace=true, % try to break only at spaces
  breakautoindent=false,
  breakindent=0pt,
  columns=fullflexible,   % optional but avoids odd gaps
  inputencoding=utf8,         % tell listings the file is UTF-8
  literate={–}{{\textendash} }1
           {"}{{\textquotedblleft}}1
           {"}{{\textquotedblright}}1
           {’}{{\textquoteright}}1 % right single quote
           {‘}{{\textquoteleft}}1 % left single quote
           {!}{{!}}1,
]
You are a professional presentation layout designer. Your task is to refine a presentation's layout by selecting appropriate visual templates from the provided options. You will optimize the visual structure while preserving the content's coherence and message integrity.

Instructions:
1. Review the provided content outline, available layouts.
2. Your job is to refine the layout selections for the presentation, optimizing how the content is visually presented.
3. Use structural layouts (such as "opening" or "ending") only when they align with the content intent and user preferences.
4. Analyze the available layouts and their media types to optimize each slide's design.
5. Preserve all content decisions from the original outline while improving the visual presentation.

For each slide in the content outline, add or modify the following fields:
- layout: Select an appropriate layout from the provided options, matching the slide's purpose and media type (mainly looking at the "concise_layout" field).

Your primary focus is to:
1. Ensure layouts are appropriate for the content and purpose of each slide
2. Maintain visual consistency throughout the presentation
3. Follow the layout_recommendation if provided, but use your judgment to select the best specific template

Note: 
- Keep layout_justification short and concise.
- Keep all original content intact - ONLY modify layout selections.

Output requirements:
- Output must be in JSON format, maintaining the exact structure of the original content outline.
- Add appropriate layout to each slide.
- Do not change any content fields (purpose, speech draft, subsections, etc.)

Example output (showing how to add layout info to existing content):
{
    "1_Opening of XX": {
        "purpose": "...", [PRESERVED FROM ORIGINAL]
        "speech_draft": "...", [PRESERVED FROM ORIGINAL]
        "subsections": [], [PRESERVED FROM ORIGINAL]
        "content_style": "...", [PRESERVED FROM ORIGINAL]
        "layout": "slide_0",
        "layout_justification": "This opening layout provides clear focus on the title while setting the presentation tone"
    },
    "2_Introduction": {
        "purpose": "...", [PRESERVED FROM ORIGINAL]
        "speech_draft": "...", [PRESERVED FROM ORIGINAL]
        "subsections": ["Title of Subsection 1.1", "Title of Subsection 1.2"], [PRESERVED FROM ORIGINAL]
        "content_style": "...", [PRESERVED FROM ORIGINAL]
        "layout": "slide_5",
        "image_assets": [PRESERVED FROM ORIGINAL]
        "layout_justification": "This layout provides adequate space for the multiple subsections while maintaining readability"
    },
    "3_XX Method": {
        "purpose": "...", [PRESERVED FROM ORIGINAL]
        "speech_draft": "...", [PRESERVED FROM ORIGINAL]
        "subsections": ["Title of Subsection 1.1", "Title of Subsection 1.2"], [PRESERVED FROM ORIGINAL]
        "content_style": "...", [PRESERVED FROM ORIGINAL]
        "layout": "slide_1",
        "image_assets": [PRESERVED FROM ORIGINAL]
        "layout_justification": "This layout balances the methodological explanation with visual reinforcement from the image"
    },
...
}

Input:
<Original Content Outline Begins>
{{ content_outline }}
<Original Content Outline Ends>

<Structural Layouts Begins>
You can only use the following layouts:
{{ functional_keys }}
<Structural Layouts Ends>

Your Output:
\end{lstlisting}
    \caption{Instruction for layout selection.}
    \label{fig:prompt_layout_selection}
\end{figure*}

\begin{figure*}[htbp]
  \centering
\begin{lstlisting}[
  literate={–}{{\textendash} }1
           {"}{{\textquotedblleft}}1
           {"}{{\textquotedblright}}1
           {’}{{\textquoteright}}1 % right single quote
           {‘}{{\textquoteleft}}1 % left single quote
           {!}{{!}}1,
]
You are an unbiased presentation analysis judge responsible for evaluating the quality of a newly generated presentation by comparing its flow and style against a document-slides sample pair. Review both presentations, focusing on structural organization, stylistic conventions (e.g., visual formatting, pacing, level of detail, and use of summaries or transitions), and the general method of delivering information. Do not consider actual topic differences between the generated presentation and the document-slides sample.

Scoring Criteria (Five-Point Scale):

1 Point (Fundamentally Dissimilar):
The generated presentation flow is structured very differently from the sample (e.g., one is linear, the other nonlinear or fragmented). Distinct presentation styles (e.g., one is highly technical and text-heavy, the other is conceptual and visual). No shared approach to organizing or delivering information.

2 Points (Minimally Similar):
Some overlap in general structure (e.g., both follow an intro-body-conclusion format) but the generated presentation differs significantly from the sample in how information is prioritized or formatted. Style elements (e.g., use of visuals, bullet points, slide density) are inconsistent. Transitions and pacing show different communication goals (e.g., one is exploratory, the other directive).

3 Points (Moderately Similar):
Both sample and generated presentations follow a comparable structure (e.g., clear sectioning, logical transitions). Share some stylistic elements (e.g., both use visuals or summaries), but execution differs. Information delivery is similar in intent but differs in granularity or emphasis.

4 Points (Strongly Similar):
Flow of the generated presentation is nearly identical to the sample, with clear parallels in section progression and use of transitions. Presentation styles are aligned (e.g., both are visual, concise, and prioritization of insights over technical depth). Minor stylistic or formatting differences, but overall experience feels consistent.

5 Points (Nearly Identical):
Generated presentation flow mirrors the sample in sequence, transitions, and internal logic. Style choices are indistinguishable in formatting, use of visuals, and pacing. Audience would perceive both presentations as stylistically and structurally the same, possibly even created by the same person or team.

Example Output:
{
  "reason": "Your short justification for the score in 2-3 sentences.",
  "score": int
}

Input:
Sample Presentation: 
{{ref_structure}}

Generated Presentation:
{{pres_structure}}

Please evaluate the slides step by step, ensuring your judgment strictly adheres to the scoring criteria. Use the full range of the scale to highlight meaningful differences. 

\end{lstlisting}
  \caption{The prompt used to evaluate the \textbf{preference-based content structure similarity} metric.}
  \label{fig:eval_pref_content_structure_prompt}
\end{figure*}

\begin{figure*}[bttp] 
    \centering
\begin{lstlisting}[
  literate={–}{{\textendash} }1
           {"}{{\textquotedblleft}}1
           {"}{{\textquotedblright}}1
           {’}{{\textquoteright}}1 % right single quote
           {‘}{{\textquoteleft}}1 % left single quote
           {!}{{!}}1,
]
You are an unbiased presentation analysis judge responsible for evaluating how closely a generated presentation adheres to a provided slide template. Review both presentations, focusing on the following aspects: Slide layout, Background design (such as colors or images) , Color scheme, Font style, Placeholder usage, Recurring elements like logos, headers, or footers.

You should ignore the specific text and image content and only evaluate the conformity to the template's design and structure.

Scoring Criteria (Five-Point Scale):

1 Point (No Adherence):
The generated presentation shows no evidence of using the template. It uses completely different background design, color palette, fonts, and slide layout.

2 Points (Minimal Adherence):
A few elements from the template are present (e.g., a primary color or font), but they are applied inconsistently. Much of the slide layout and background design does not match the template's master slides, and core branding elements (like logos or footers) are missing or misplaced.

3 Points (Moderate Adherence):
The generated presentation uses the template's general background, color scheme, and font style, but with some inconsistencies. Some slides use the correct master layout, while others deviate significantly. The overall impression is a partially successful attempt to follow the template.

4 Points (Strong Adherence):
The generated presentation correctly uses most of the template's features, including background design, color palette, fonts, and the most common slide layout (e.g., title, content, section header). There may be minor deviations, such as incorrect use of a less common layout or slight inconsistencies in formatting.

5 Points (Strict Adherence / Nearly Identical):
The generated presentation perfectly adheres to the template. All slides correctly use the intended master layout, background design, color palette, fonts, and placeholders. The presentation looks professionally and consistently designed according to the template's rules.

Input:
{{num_of_target_slide}} pages of slide_images and {{num_of_template_slide}} pages of template_images.

Example Output:
{
  "reason": "Your short justification for the score in 2-3 sentences.",
  "score": int
}

You only need to give a unified `reason` and `score` for all provided slide_images and template_images. Do not generate any other text except from the json containing `reason` and `score`.

Please evaluate the slides step by step, ensuring your judgment strictly adheres to the scoring criteria. Be strict and use the full range of the scale to highlight meaningful differences. 
\end{lstlisting}
    \caption{The prompt used to evaluate the \textbf{preference-based aesthetic similarity} metric.}
    \label{fig:eval_pref_aesthetic_prompt}
\end{figure*}

\begin{figure*}[htbp] 
    \centering
\begin{lstlisting}[
  literate={–}{{\textendash} }1
           {"}{{\textquotedblleft}}1
           {"}{{\textquotedblright}}1
           {’}{{\textquoteright}}1 % right single quote
           {‘}{{\textquoteleft}}1 % left single quote
           {!}{{!}}1,
]
You are an unbiased presentation analysis judge responsible for evaluating the content of a set of slides. Your evaluation should consider both the informativeness of the slides in relation to the target paper, and the quality of the presentation content itself (e.g., clarity).

Scoring Criteria (Five-Point Scale):

1 Point (Poor):
The slides contain no meaningful content from the paper or misrepresent it. Text has significant errors, is poorly structured, or visuals are distracting/irrelevant. The slides are difficult to understand.

2 Points (Below Average):
The slides contain only scattered, low-impact points from the paper. It fails to convey any core concepts. Lacks a clear focus, text is awkwardly phrased, or visuals are weak and do not support the content.

3 Points (Average):
The slides present a general idea or some minor details from the paper but may lack depth, context, or key technical substance. The content is understandable but may lack visual appeal or could be organized more effectively. Text and images are not well-integrated.

4 Points (Good):
The slides clearly present a major component of the paper (e.g., motivation, a key part of the method, a result) with reasonable detail. The slides are well-developed and clear. Visuals are relevant and support the text, with only minor room for improvement in design.

5 Points (Excellent):
The slides comprehensively and accurately cover a core contribution, methodology, or result from the paper with appropriate depth. The slides are exceptionally well-structured with a clear focus. Text is concise and polished. Images and text are highly synergistic and effectively convey the information.

Example Output:
{
  "reason": "Your short justification for the score in 1-2 sentences.",
  "score": int
}

Input:
<Research Paper Begins>
{{paper}}
<Research Paper Begins>

Please evaluate the slides step by step, ensuring your judgment strictly adheres to the scoring criteria. Be strict and use the full range of the scale to highlight meaningful differences. 
\end{lstlisting}
    \caption{The prompt used to evaluate the \textbf{preference-independent content} metric.}
    \label{fig:eval_content_structure_prompt}
\end{figure*}

\begin{figure*}[htbp] 
    \centering
\begin{lstlisting}[
  literate={–}{{\textendash} }1
           {"}{{\textquotedblleft}}1
           {"}{{\textquotedblright}}1
           {’}{{\textquoteright}}1 % right single quote
           {‘}{{\textquoteleft}}1 % left single quote
           {!}{{!}}1,
]
You are an unbiased presentation analysis judge responsible for evaluating the visual appeal of a set of slides. Please carefully review the provided description of the slides, assessing their aesthetics only, and provide your judgment score. 

Scoring Criteria (Five-point scale):

1 Point (Poor):
The slides have poor color scheme and no visual appeal, to the extent of making the content difficult to read.

2 Points (Below Average):
The slides use monotonous colors (black and white), ensuring readability while lacking visual appeal.

3 Points (Average):
The slides employ a basic color scheme; however, they lack supplementary visual elements such as icons, backgrounds, images, or geometric shapes (like rectangles), making it look plain.

4 Points (Good):
The slides use a harmonious color scheme and contain some visual elements (like icons, backgrounds, images, or geometric shapes); however, minor flaws may exist in the overall design.

5 Points (Excellent):
The style of the slides is harmonious and engaging. The use of supplementary visual elements like images and geometric shapes enhances the slides' overall visual appeal.


Example Output:
{
  "reason": "Your short justification for the score in 1-2 sentences.",
  "score": int
}

Input: {{descr}}

Please evaluate the slides step by step, ensuring your judgment strictly adheres to the scoring criteria. Be strict and use the full range of the scale to highlight meaningful differences. 
\end{lstlisting}
    \caption{The prompt used to evaluate the \textbf{preference-independent aesthetic} metric.}
    \label{fig:eval_aesthetic_prompt}
\end{figure*}

\begin{figure*}[htbp]
  \centering
\begin{lstlisting}
Task: Document-to-Slide Presentation Generation

You are a professional slideshow creator. You are tasked with transforming an academic document/paper into a high-quality PowerPoint presentation This task involves several key components:

Input:
- A target document (PDF) that you need to convert into slides (the pdf file in one of 1-150.pdf)
- A reference document (PDF) and its corresponding presentation (PPT) that serve as style and format guidance (the .pdf pairs with the same name)
- A PowerPoint template to use for the new presentation (the .pptx file)

Parameters:
- Number of slides to generate: 10

The presentation should follow the style, tone, and formatting conventions seen in the reference presentation.

Output: A complete 10-slide presentation in PowerPoint format that effectively communicates the key content of the target document while following the style and format of the reference presentation.

\end{lstlisting}
  \caption{The prompt template for generating slides with ChatGPT.}
  \label{fig:chatgpt_prompt}
\end{figure*}

\begin{figure*}[htbp]
  \centering
\begin{lstlisting}
Generate a 10-slide presentation for the following document:
{{target_doc_source}}

You should follow the sample document (PDF) and its corresponding presentation (PPT) that serve as style and format guidance.
Sample document:
{{sample_doc_source}}

Sample presentation:
{{sample_presentation_source}}

Now, complete 10-slide presentation that effectively communicates the key content of the target document while following the style and format of the sample presentation.
\end{lstlisting}
  \caption{The prompt template for generating slides with AutoPresent.}
  \label{fig:autopresent_prompt}
\end{figure*}

\end{document}